\documentclass[letterpaper, 10 pt]{IEEEtran}

\usepackage[numbers,sort&compress]{natbib}
\usepackage[caption=false]{subfig}
\usepackage{hhline}

\usepackage{amsfonts}
\usepackage{amsmath}
\usepackage{amssymb}
\usepackage{amsthm}
\usepackage{graphicx}

\usepackage[noend]{algorithmic}
\usepackage{multirow}
\usepackage{xcolor}
\usepackage{prettyref}
\usepackage{xspace}
\usepackage{bm}
\usepackage{tikz}

\usepackage{hyperref} %
\newcommand{\name}{CAST$^\star$\xspace}
\newcommand{\nameSharp}{CAST\#\xspace}
\newcommand{\nameRaw}{CAST\xspace}

\newcommand{\Bp}{\mathbf{p}}
\newcommand{\Bv}{\mathbf{v}}
\newcommand{\BR}{\mathbf{R}}
\newcommand{\BOmega}{\boldsymbol{\Omega}}

\newrefformat{appendix}{Appendix\,\ref{#1}}

\newrefformat{prob}{Problem\,\ref{#1}}
\newrefformat{def}{Definition\,\ref{#1}}
\newrefformat{sec}{Section\,\ref{#1}}
\newrefformat{sub}{Section\,\ref{#1}}
\newrefformat{prop}{Proposition\,\ref{#1}}
\newrefformat{app}{Appendix\,\ref{#1}}
\newrefformat{alg}{Algorithm\,\ref{#1}}
\newrefformat{cor}{Corollary\,\ref{#1}}
\newrefformat{thm}{Theorem\,\ref{#1}}
\newrefformat{lem}{Lemma\,\ref{#1}}
\newrefformat{fig}{Fig.\,\ref{#1}}
\newrefformat{tab}{Table\,\ref{#1}}

\newtheorem{theorem}{Theorem}
\newtheorem{problem}{Problem}

\newtheorem{corollary}[theorem]{Corollary}

\newtheorem{proposition}[theorem]{Proposition}

\newcommand{\bdmath}{\begin{dmath}}
\newcommand{\edmath}{\end{dmath}}
\newcommand{\beq}{\begin{equation}}
\newcommand{\eeq}{\end{equation}}
\newcommand{\bdm}{\begin{displaymath}}
\newcommand{\edm}{\end{displaymath}}
\newcommand{\bea}{\begin{eqnarray}}
\newcommand{\eea}{\end{eqnarray}}
\newcommand{\beal}{\beq \begin{array}{ll}}
\newcommand{\eeal}{\end{array} \eeq}
\newcommand{\beas}{\begin{eqnarray*}}
\newcommand{\eeas}{\end{eqnarray*}}
\newcommand{\ba}{\begin{array}}
\newcommand{\ea}{\end{array}}
\newcommand{\bit}{\begin{itemize}}
\newcommand{\eit}{\end{itemize}}
\newcommand{\ben}{\begin{enumerate}}
\newcommand{\een}{\end{enumerate}}

\newcommand{\eg}{\emph{e.g.,}\xspace}
\newcommand{\ie}{\emph{i.e.,}\xspace}

\renewcommand{\boldsymbol}[1]{{\bm #1}}

\newcommand{\hide}[1]{}

\newcommand{\hiddenText}{{\color{gray} hidden text.}}
\newcommand{\hideWithText}[1]{\hiddenText}

\DeclareMathOperator*{\argmax}{arg\,max}

\newcommand{\eye}{{\mathbf I}}

\newcommand{\SO}[1]{\ensuremath{\mathrm{SO}(#1)}\xspace}

\newcommand{\vv}{\boldsymbol{v}}

\newcommand{\vxx}{\boldsymbol{x}}

\newcommand{\blue}[1]{{\color{blue}#1}}

\newcommand{\linkToPdf}[1]{\href{#1}{\blue{(pdf)}}}
\newcommand{\linkToPpt}[1]{\href{#1}{\blue{(ppt)}}}
\newcommand{\linkToCode}[1]{\href{#1}{\blue{(code)}}}
\newcommand{\linkToWeb}[1]{\href{#1}{\blue{(web)}}}
\newcommand{\linkToVideo}[1]{\href{#1}{\blue{(video)}}}
\newcommand{\linkToMedia}[1]{\href{#1}{\blue{(media)}}}
\newcommand{\award}[1]{\xspace} %

\begin{document}

\newcommand{\minproblem}[3]{
    \min_{#1} \: & #2 \\
    \textrm{s.t.} \quad & #3
}
\renewcommand{\baselinestretch}{0.94}

\title{A Certifiable Algorithm for Simultaneous \\ Shape Estimation and Object Tracking
}

\author{\authorblockN{Lorenzo Shaikewitz, Samuel Ubellacker, Luca Carlone}
        \authorblockA{Laboratory for Information \& Decision Systems (LIDS)\\
              Massachusetts Institute of Technology\\
              Cambridge, USA\\
              Email: \{lorenzos, subella, lcarlone\}@mit.edu}}
\author{Lorenzo Shaikewitz, Samuel Ubellacker, and Luca Carlone
    \thanks{
    This work was supported by the AFOSR "Certifiable and Self-Supervised Category-Level Tracking" program, Carlone's NSF CAREER award, the ONR RAPID program, and by MathWorks.}
    \thanks{All authors are with the Laboratory for Information and Decision Systems, Massachusetts Institute of Technology, Cambridge, MA.
            {\tt\footnotesize \{lorenzos, subella, lcarlone\}@mit.edu}}
    }

\maketitle

\begin{tikzpicture}[overlay, remember picture]
\path (current page.north east) ++(-3.7,-0.2) node[below left] {
\textbf{This paper has been accepted for publication in \emph{IEEE Robotics and Automation Letters.}}
};
\end{tikzpicture}
\begin{tikzpicture}[overlay, remember picture]
\path (current page.north east) ++(-4.3,-0.6) node[below left] {
Please cite the paper as: Lorenzo Shaikewitz, Samuel Ubellacker, and Luca Carlone,
};
\end{tikzpicture}
\begin{tikzpicture}[overlay, remember picture]
\path (current page.north east) ++(-4.3,-1) node[below left] {
``A Certifiable Algorithm for Simultaneous Shape Estimation and Object Tracking''
};
\end{tikzpicture}
\begin{tikzpicture}[overlay, remember picture]
\path (current page.north east) ++(-6.7,-1.4) node[below left] {
    \emph{IEEE Robotics and Automation Letters}, Dec. 2024.
};
\end{tikzpicture}

\begin{abstract}
    Applications from manipulation to autonomous vehicles rely on robust and general object tracking to safely perform  tasks in dynamic environments. We propose the first certifiably optimal category-level approach for simultaneous shape estimation and pose tracking of an object of known category (\eg a car). Our approach uses 3D semantic keypoint measurements extracted from an RGB-D image sequence, and phrases the estimation as a fixed-lag smoothing problem.
    Temporal constraints enforce the object's rigidity (fixed shape) and smooth motion according to a constant-twist motion model. The solutions to this problem are the estimates of the object's state (poses, velocities) and shape (paramaterized according to the \textit{active shape model}) over the smoothing horizon. %
    Our key contribution is to show that despite the non-convexity of the fixed-lag smoothing problem, we can solve it to \emph{certifiable optimality} using a small-size semidefinite relaxation. 
    We also present a 
    fast outlier rejection scheme that filters out incorrect keypoint detections with shape and time compatibility tests, and
    wrap our certifiable solver in a graduated non-convexity scheme.
    We evaluate the proposed approach on synthetic and real data, showcasing its performance in a table-top manipulation scenario and a drone-based vehicle tracking application.
\end{abstract}

\vspace{-10pt}
\section{Introduction}
\label{sec:introduction}

{Target} 
tracking is a well-studied problem across multiple research communities, including robotics, computer vision, and aerospace.
Early work models the target as a point mass and is concerned with estimating its location from measurements (\eg bearing vectors) while resolving data association, \eg~\cite{Fortmann80}.
In modern robotics applications, robots navigate in close proximity and possibly interact with nearby objects. 
For those applications, the robot also needs to estimate the object's {shape}. This leads to the coupled problem of \textit{shape estimation and pose tracking}, which is crucial for autonomous vehicles~\cite{peng23-trackingDriving}, table-top manipulation~\cite{Wen20iros-seTrack}, monitoring and surveillance~\cite{Lepetit05-monocularTracking}, among other applications.

A significant body of work is dedicated to object pose and shape estimation in the \emph{single-frame case}~\cite{Bruns24ras-posesurvey, Shi23tro-PACE, Wang19-normalizedCoordinate}. However, using a single image for detection and estimation sacrifices temporal information which is readily available. A good tracking algorithm leverages temporal consistency to operate in noisy, occlusion-rich, or highly dynamic environments.
When the object shape is known exactly, \emph{instance-level} tracking algorithms leverage known geometry~\cite{Yang20tro-teaser, Lepetit05-monocularTracking} or extensive training on the specific object instance~\cite{Wen20iros-seTrack, Wang23iccv-deepac,Deng2019rss-poserbpf}. However, in practical applications the object is rarely known exactly. Autonomous vehicles must track any of the thousands of car models in their surroundings; even if all were to be cataloged, the vehicle must still reason over intra-category variations.

\begin{figure}[tb]
\centerline{\includegraphics[width=\linewidth]{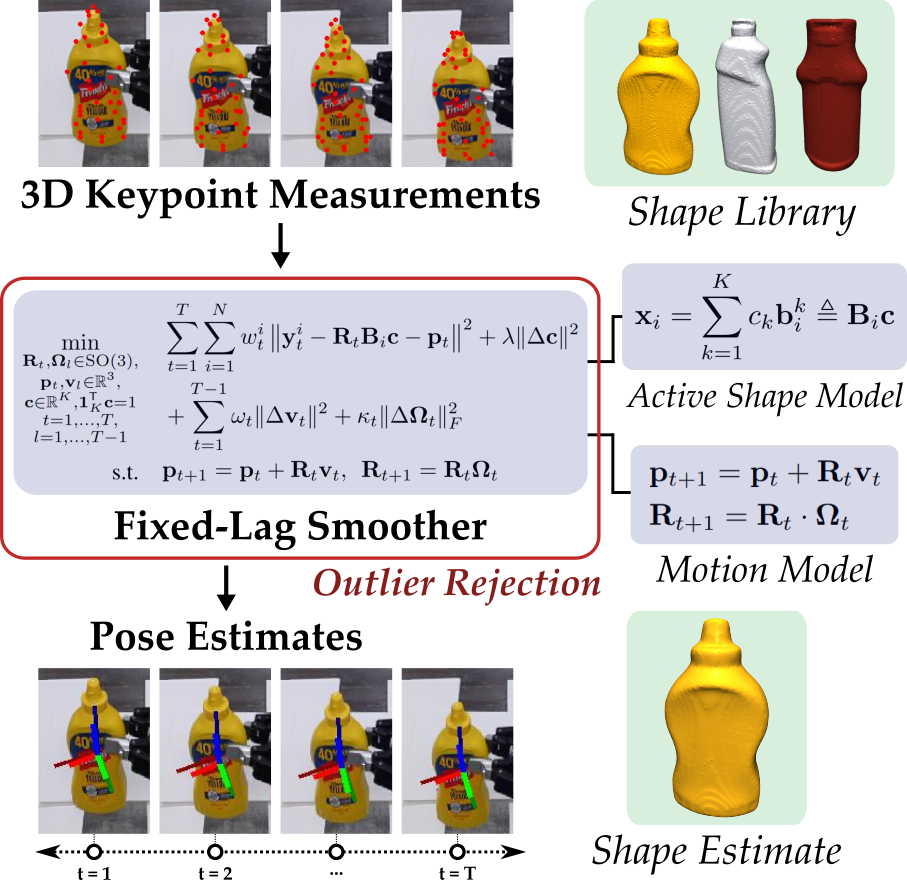}}
\caption{\textbf{Overview of \nameSharp}. {We estimate the shape and track the pose of an object from a sequence of images picturing the object. Given 3D keypoint measurements obtained via a learning-based detector, we formulate a non-convex fixed-lag smoothing problem 
where the shape is parametrized using an active shape model and motion smoothness is enforced using a constant-twist motion model. %
We solve this problem via a tight and small-size semidefinite relaxation and wrap the method in an outlier rejection scheme to robustly estimate shape and pose over a fixed time horizon.}}
\label{fig:intro}
\end{figure}

Recent work has made considerable progress towards \emph{category-level} and \emph{category-free} object tracking using end-to-end learning~\cite{Weng21iccv-captra, Sun22iros-icktrack} or learning-based segmentation combined with local solvers for estimation~\cite{Wang20icra-6pack, Wen23cvpr-bundlesdf,Runz18ismar-maskfusion}. While often effective in practice, learning-based approaches have unpredictable failure modes when used outside their training domain, while 
local-solver-based approaches are prone to converge to local minima corresponding to poor estimates.
Safety-critical applications demand interpretable and predictable models that produce provably optimal estimates. 

{\bf Contribution.}
We propose the first certifiably optimal algorithm for category-level shape estimation and pose tracking (Fig.~\ref{fig:intro}). 
In our problem formulation (Section~\ref{sec:problem}), we consider as input 
the 3D position of keypoints detected in an RGB-D image sequence.
These inputs are fed to a fixed-lag smoother that performs estimation of the states over a receding horizon, while enforcing smoothness of the motion using a constant-twist motion model. 
We parametrize the object's shape using an \textit{active shape model}~\cite{Cootes95cviu,Zhou15cvpr}, which describes shape as a linear combination of 3D models from a library of representative models. 
Our first key contribution (Section~\ref{sec:method}) is to show that despite the non-convexity of the estimation problem, 
we can develop an empirically tight semidefinite relaxation that computes certifiably optimal object poses, velocities, and shape without the need for an initial guess. We name the resulting approach \emph{\name} (\emph{Certifiable Algorithm for Shape estimation and Tracking}).
Our second contribution (Section~\ref{sec:outlier}) is to extend \emph{\name} to the practical case where some of the keypoint measurements are outliers; we handle this case by combining an outlier-pruning method with a robust estimator and graduated non-convexity~\cite{Yang20ral-GNC}. We call the resulting outlier-robust approach \emph{\nameSharp}.
 Our final contribution (Section~\ref{sec:experiments}) is to evaluate our methods 
 in extensive experiments, and show they produce accurate estimates when tested in simulation, 
 on the YCBInEOAT~\cite{Wen20iros-seTrack} and NOCS~\cite{Wang19-normalizedCoordinate} datasets, and in a drone-based vehicle tracking scenario.

\section{Related Work}
\textbf{Instance-Level Object Tracking.}
Traditional target tracking approaches circumvent shape estimation by assuming the object to be a point mass~\cite{BarShalom92} or assuming full knowledge of the object shape~\cite{Lepetit05-monocularTracking,Blackman99-modernTracking,Yang20tro-teaser, Deng2019rss-poserbpf, Piga2021frontiers-maskukf, Wen20iros-seTrack, Wang23iccv-deepac}. 
{
Early approaches used handcrafted features, such as points, edges~\cite{Simon98iccv-tracking}, or planes~\cite{Simon02-planarPose} to compute relative poses. The set of pose estimates could then be smoothed via Kalman filtering~\cite{Lepetit05-monocularTracking,Blackman99-modernTracking}. 
}
More recently, the use of handcrafted features has given way to learned features~\cite{Wen20iros-seTrack} and edge detection~\cite{Wang23iccv-deepac}, and new approaches based on point cloud registration~\cite{Yang20tro-teaser}, particle filtering~\cite{Deng2019rss-poserbpf}, or unscented Kalman filters~\cite{Piga2021frontiers-maskukf} have emerged.

\textbf{Category-Level and Category-Free Object Tracking.} %
In practical settings, %
instance-level information is rarely available. Recent approaches investigate pose and shape estimation for objects within a known category~\cite{Wang20icra-6pack, Wen21iros-bundletrack, Weng21iccv-captra, Runz18ismar-maskfusion, Sun22iros-icktrack} or at least similar enough to the training data~\cite{Wen23cvpr-bundlesdf}. 
These approaches generally extract a sparse representation of the object to estimate relative motion between frames. Wang et al.~\cite{Wang20icra-6pack} focus on an attention mechanism for extracting frame-to-frame keypoints in a self-supervised manner, leaving the work of relative pose estimation to point cloud registration, which is 
unable to use temporal information beyond two frames. Wen and Bekris~\cite{Wen21iros-bundletrack} use a similar architecture but take a SLAM-inspired approach, using dense frame-to-frame feature correspondences and multi-frame pose graph optimization to refine the estimate. 
{
    Other methods use learned keypoint correspondences for the Iterative Closest Point (ICP) method~\cite{Sun22iros-icktrack} or learning-based regression to estimate relative motion in the small pose regime~\cite{Weng21iccv-captra}.
} 
Even with keyframe selection~\cite{Wen23cvpr-bundlesdf}, frame-to-frame back-ends require a separate tool to obtain object pose relative to a camera or world frame, which is often useful in applications.
In contrast, we propose an optimization back-end that produces \textit{certifiably optimal} shape and pose estimates from category-level keypoints without relying on local solvers. This gives useful world-frame poses directly and allows the use of a motion model to mitigate the impact of measurement noise.

\textbf{Certifiable Algorithms.}
Our work extends the body of work on~\emph{certifiable perception algorithms}: a certifiable algorithm solves an optimization problem and either provides a certificate of optimality or a bound on the suboptimality of the produced solution~\cite{Yang20tro-teaser}. Certifiable algorithms are typically derived using semidefinite relaxations, and are usually based on Shor's relaxation of Quadratically Constrained Quadratic Programs (QCQP) or Lasserre's relaxation of polynomial optimization problems~\cite{Yang22pami-certifiablePerception,Lasserre10book-momentsOpt, Shor87}. Certifiable algorithms have been proposed for rotation averaging~\cite{Brynte21-semidefiniteRotation, Saunderson14cdc},
pose graph optimization~\cite{Rosen18ijrr-sesync,Carlone15icra-verification}, 
3D registration~\cite{Briales17cvpr-registration,Yang20tro-teaser},  
2-view geometry~\cite{Zhao20cvpr-certifiablyEssential,Garcia21IVC-certifiablerelativepose},
perspective-n-point problems~\cite{Sun20access-certifiablyPnP},
and single-frame pose and shape estimation~\cite{Shi23tro-PACE}.
Recent work has extended certifiable solvers to cope with outliers~\cite{Yang22pami-certifiablePerception} 
and anisotropic noise~\cite{Holmes24-semidefinite}.
Our approach extends~\cite{Shi23tro-PACE} to tracking over a receding time horizon using a motion model.

\section{Problem Formulation}
\label{sec:problem}
This section formalizes the \textit{category-level shape estimation and pose tracking} problem. 
Given a sequence of RGB-D images picturing an object of known category (\eg a car), and assuming 
the availability of a 3D semantic keypoint detector, we seek an estimate of the time-independent shape and time-dependent pose (\ie position and orientation) of the object. Below we describe our choice of motion model, shape representation, and measurement model.

\textbf{Object State and Motion Model.}
We represent the target object's state using its pose and velocity at a particular time $t$. Denote the position and orientation of the target object in the world frame as $\Bp_t\in\mathbb{R}^3$ and $\BR_t\in\SO{3}$, respectively. Similarly, denote the target's \emph{body-frame} change in position and change in rotation between time steps with $\Bv_t\in\mathbb{R}^3$ and $\BOmega_t\in\SO{3}$. These state variables are the discrete time analog to velocity and rotation rate. Any object's motion obeys the following discrete-time first-order dynamics:
\begin{equation}
    \label{eq:prUpdate}
    \Bp_{t+1} = \Bp_t + \BR_t\Bv_t,\quad
    \BR_{t+1} = \BR_t \cdot \BOmega_t
\end{equation}
The model~\eqref{eq:prUpdate} is quite general, since by choosing suitable values of $\Bv_t$, $\BOmega_t$ we can produce arbitrary trajectories. %

Now, we assume that the velocities' dynamics are approximately \emph{constant twist}; \ie the body-frame velocity and the rotation rate are constant during short time intervals up to random perturbations $\vv^{\epsilon}_t \in \mathbb{R}^3$ and $\BR^{\epsilon}_t\in\SO{3}$:
\begin{equation}
    \label{eq:velUpdate}
        \Bv_{t+1} = \Bv_t + \vv^{\epsilon}_t,\quad
        \BOmega_{t+1} =  \BOmega_t \cdot \BR^{\epsilon}_t
\end{equation}
When $\Bv_t$ and $\BOmega_t$ are exactly constant the dynamical system in equations~\eqref{eq:prUpdate}-\eqref{eq:velUpdate} models 3D spiral-shaped trajectories, including the corner cases of a straight line, circular trajectory, or in-place rotation. The random noise terms model small deviations from these assumptions in the observed trajectory.
The proposed constant-twist model is a 3D version of the popular constant-turn-rate model~\cite{Blackman99-modernTracking}, generalizing it to allow an arbitrary axis of rotation. Such a model is expressive enough to capture the non-holonomic motion of a car and the unpredictable motions of a manipulated object.

In the following, we assume that the velocity noise follows an isotropic zero-mean Gaussian distribution: $\vv_t^\epsilon\sim\mathcal{N}(0,\boldsymbol{\Sigma}_t^v)$ and that the relative rotation noise follows an isotropic Langevin distribution about the identity, following standard practice~\cite{Rosen18ijrr-sesync} for distributions over $\SO{3}$:
$\BR^{\epsilon}_t\sim\mathcal{L}(\mathbf{I}_3, \kappa_t)$. In this equation, $\kappa_t$ is the \emph{concentration parameter} of the Langevin distribution (intuitively, this plays a similar role as the inverse of the variance).

\textbf{Shape Parameterization}
\begin{figure}[tb]
\centerline{\includegraphics[width=0.8\linewidth]{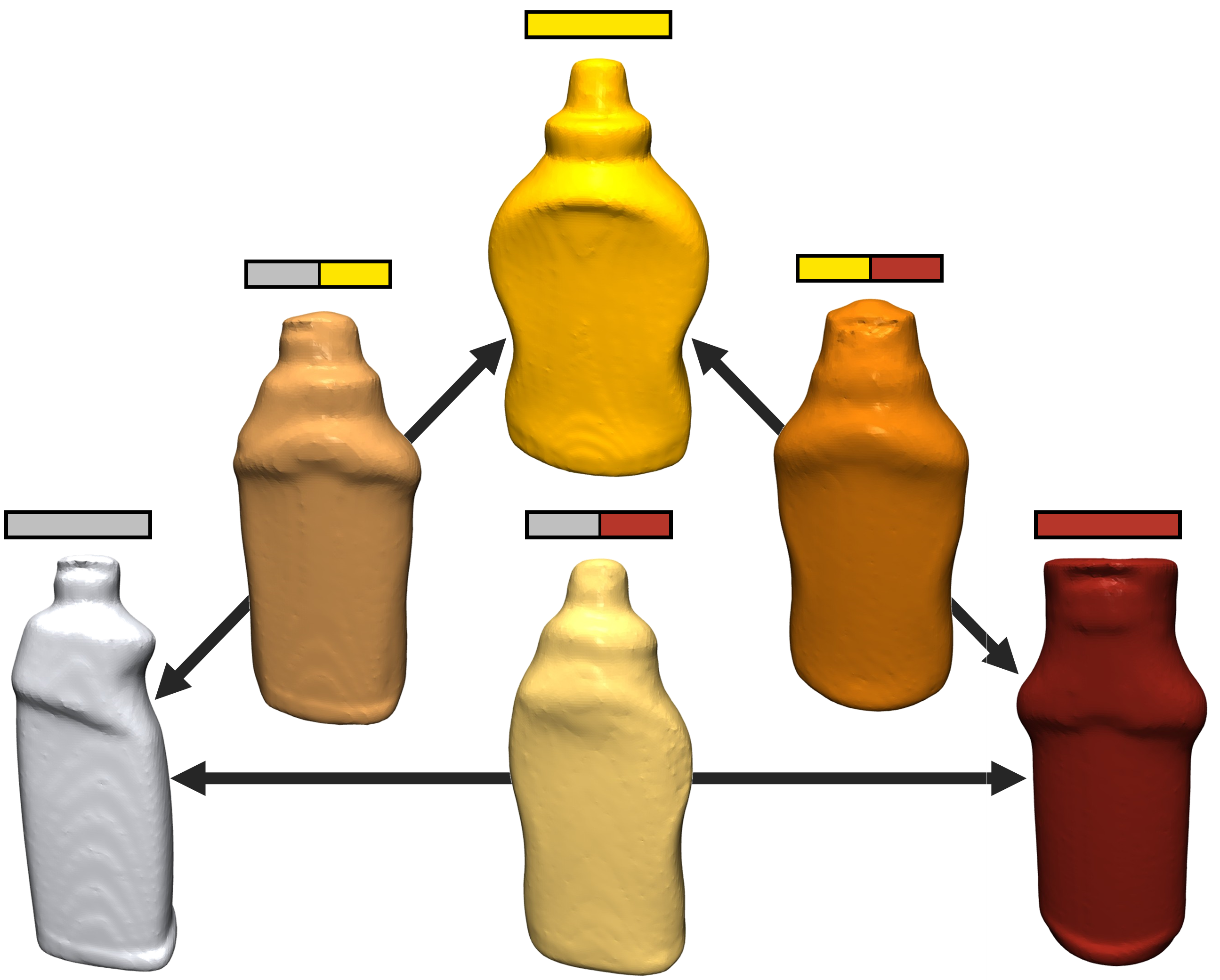}}
\caption{\textbf{Active Shape Model.} Known 3D models in the \emph{bottle} category and their averages computed according to the active shape model. Vertices are the original models and edges are the average of the two vertices. The active shape model can represent any 3D geometry in the convex hull of its shape library through a point-wise weighted average.}
\label{fig:asm}
\end{figure}
We use the \textit{active shape model} to represent intra-category shape variations. Given an object 
 category (\eg\hspace{0.5pt}{bottle}), we assume a library of 3D models (\eg\hspace{0.5pt}{specific bottle shapes}) that span the category, 
 where the objects in the library are denoted as $\mathcal{B}_k$, $k=1,\hdots,K$. Any instance, then, is just a pointwise linear combination of the models in the shape library (see \prettyref{fig:asm}). More formally, let $\mathbf{x}_i$ be a point on the instance object corresponding to the point $\mathbf{b}_i^k\in\mathcal{B}_k$ in each library shape. The active shape model is:
\begin{equation}
    \label{eq:activeshapemodel}
    \mathbf{x}_i = \sum_{k=1}^K c_k \mathbf{b}_i^k \triangleq \mathbf{B}_i \mathbf{c}
\end{equation}
where $c_k\in[0,1]$ and $\sum_k c_k = 1$. Thus, the shape of the target object is fully specified by its shape coefficient $\mathbf{c}=[c_1,\hdots,c_K]$ and the shape library for each point $\mathbf{B}_i=[\mathbf{b}_i^1,\hdots,\mathbf{b}_i^K]$. This representation is simple and expressive: it captures any object in the convex hull of the shape library (including the library shapes themselves) via a linear combination described by a single vector of coefficients~\cite{Cootes95cviu,Zhou15cvpr}.
 Further, measurements of a small number of semantic \emph{keypoints} are enough to resolve the dense object shape.

\textbf{Measurement Model.}
The inputs to our estimator are measurements of the 3D positions of \emph{semantic keypoints} on the target object. These keypoints correspond to semantically meaningful features common to a specific object category, and are typically produced by a learning-based detector, as in \cite{Shi23tro-PACE, Pavlakos17icra-semanticKeypoints}. For instance, a set of keypoints on a bottle might be the locations of the cap, center-point of the base, label, etc. Such keypoints may be detected by a model trained on a category of bottles, not just a particular instance.

At each time $t$ we are given the 3D position of $N$ keypoints denoted $\mathbf{y}_t^1,\hdots,\mathbf{y}_t^N$. These measurements obey the following generative model:
\begin{equation}
    \label{eq:measurements}
    \mathbf{y}_t^i = \BR_t \cdot (\mathbf{B}_i \mathbf{c}) + \Bp_t + \boldsymbol{\epsilon}_t^i
\end{equation}
Each measurement $\mathbf{y}_t^i$ is a rigid transformation $(\BR_t, \Bp_t)$ of the keypoint's location in the object's frame $\mathbf{B}_i \mathbf{c}$ (expressed according to the active shape model) plus measurement noise $\boldsymbol{\epsilon}_t$. 
For now, we assume the measurement noise obeys an isotropic zero-mean Gaussian distribution:
$\boldsymbol{\epsilon}_t^i\sim\mathcal{N}(0,\boldsymbol{\Sigma}_t^i)$.

\textbf{Simultaneous Shape Estimation and Tracking.}
We are now ready to state the problem we tackle in this paper.
\begin{problem}
    \label{prob:tracking}
    Consider an object of known category moving according to the dynamics in eqs.~\eqref{eq:prUpdate}-\eqref{eq:velUpdate}. 
    Given measurements of $N$ keypoints in the form \eqref{eq:measurements} taken over $T$ time steps, estimate the time-varying state $(\BR_t, \Bp_t, \Bv_t, \BOmega_t)$ and time-independent shape $\mathbf{c}$ of the object for $t=1,\hdots,T$.
\end{problem}

\prettyref{prob:tracking} may be interpreted as a fixed-lag smoother, where our primary goal is to estimate the state at time $T$ using also the most recent $T-1$ measurements. %

\section{\name: Certifiable Shape Estimation and Tracking in the Outlier-Free Setting}
\label{sec:method}
This section presents \name, a certifiably optimal estimator solving \prettyref{prob:tracking} in the outlier-free setting. \name is also the basis for our outlier-robust extension in \prettyref{sec:outlier}.

We adopt a \emph{maximum a posteriori} estimation framework that represents \prettyref{prob:tracking} as an optimization problem. This framework minimizes the residual errors of the measurement and motion models over the time horizon $T$, possibly including priors. In our case, the only prior is that shape coefficients $\mathbf{c}$ are distributed according to a Gaussian with covariance $\frac{1}{\lambda}\mathbf{I}_3$ about the mean shape $\bar{\mathbf{c}} \triangleq \frac{1}{K}\mathbf{1}_K$. In practice, this prior regularizes the problem when the shape library is larger than the number of keypoints ($K > N$); see e.g.~\cite{Shi23tro-PACE}.

The maximum a posteriori estimator takes the form:

\begin{equation}
    \label{eq:optprob_original}
    \begin{aligned}[c]
        \min_{
            \substack{
                \BR_t,\BOmega_l\in \SO{3}, \\
                \Bp_t,\Bv_l\in\mathbb{R}^{3},\\
                \mathbf{c}\in\mathbb{R}^K,\mathbf{1}_K^\mathsf{T}\mathbf{c}=1\\
                t = 1,\hdots,T,\\
                l = 1,\hdots,T-1
                }
        }
        \\[17pt]
        \textrm{s.t.} & 
    \end{aligned}
    \begin{aligned}[c]
        &\sum_{t=1}^T \sum_{i=1}^N w_t^i \left\| \mathbf{y}_t^i - \BR_t\mathbf{B}_i\mathbf{c} - \mathbf{p}_t\right\|^2 
        + \lambda\|\Delta\mathbf{c}\|^2
        &&\\
        &+ \sum_{t=1}^{T-1} \omega_t\|\Bv_{t+1}-\Bv_t\|^2
        + \kappa_t\|\BOmega_{t+1}-\BOmega_t\|^2_F &&\\
        &\mathbf{p}_{t+1} = \mathbf{p}_t + \mathbf{R}_t\mathbf{v}_t,\:\:
        \mathbf{R}_{t+1} = \mathbf{R}_t\boldsymbol{\Omega}_t&&
    \end{aligned}
\end{equation}
In the previous expession, we used the shorthand $\Delta\mathbf{c}\triangleq\mathbf{c}-\mathbf{\bar{c}}$ and assumed
isotropic covariances $\boldsymbol{\Sigma}_t^i \triangleq \frac{1}{w_t^i} \eye_3$ and $\boldsymbol{\Sigma}_t^v \triangleq \frac{1}{w_t} \eye_3$. We also relaxed the constraint $c_k\geq0$. We observe that the objective is the sum of the shape prior with the negative log-likelihoods of the measurements~\eqref{eq:measurements} and dynamics~\eqref{eq:velUpdate}. The constraints enforce the domains of the variables (e.g. $\mathbf{R}_t\in \SO{3}$ or $\mathbf{1}_K^\mathsf{T}\mathbf{c}=1$) and the 
 dynamics~\eqref{eq:prUpdate}. Eq. \eqref{eq:optprob_original} is a maximum a posteriori estimator; see~\prettyref{appendix:optprob} for proof.

Notice that~\eqref{eq:optprob_original} is non-convex due to the constraint set $\SO{3}$ and the quadratic equality constraints. Thus, local search methods such as gradient descent or Gauss-Newton are prone to local minima that result in bad  estimates.

In the following we present our approach to solving~\eqref{eq:optprob_original} to certifiable optimality via a semidefinite relaxation. In \prettyref{sec:shape} we simplify the problem by analytically solving for the optimal shape coefficient. Using a change of variables, we rewrite~\eqref{eq:optprob_original} as a non-convex quadratically constrained quadratic program (QCQP) in \prettyref{sec:qcqp} and apply a semidefinite relaxation in \prettyref{sec:semidefin}. This relaxed problem can be solved using traditional convex optimization techniques and is shown to be empirically \textit{tight} (\ie the relaxation solves~\eqref{eq:optprob_original} to optimality) in \prettyref{sec:experiments}.

\subsection{Closed-Form Solution for Shape}
\label{sec:shape}
Observe that~\eqref{eq:optprob_original} is a linearly constrained convex quadratic program in the variable $\mathbf{c}$. Thus, we can solve for the optimal shape coefficient $\mathbf{c^\star}$ in closed form as a function of the other unknown variables. We formalize this observation below.

\begin{proposition}[Optimal Shape]
    \label{prop:shapesoln}
    For any positions and rotations $(\mathbf{p}_t, \mathbf{R}_t)$, the optimal shape coefficient solving \eqref{eq:optprob_original} is
    \begin{equation}
        \label{eq:optshape}
        \mathbf{c^\star} = 2\mathbf{G}\left(
            \mathbf{B}^\mathsf{T}
            \sum_{t=1}^T 
            \mathbf{W}_t
            \begin{bmatrix}
                \mathbf{R}^\mathsf{T}_t ( \mathbf{y}^1_t - \mathbf{p}_t ) \\
                \vdots \\
                \mathbf{R}^\mathsf{T}_t (\mathbf{y}^N_t - \mathbf{p}_t )
            \end{bmatrix}
            + \lambda \mathbf{\bar{c}}\right) + \mathbf{g}
    \end{equation}
    where we defined the following symbols:
    \begin{equation}
    \begin{array}{lll}
    \mathbf{W}_t &\triangleq \mathrm{blkdiag}(w_1^t\mathbf{I}_3,\hdots,w_N^t\mathbf{I}_3) 
    &\in \mathbb{R}^{3N\times 3N}
    \\
    \mathbf{B} &\triangleq [\mathbf{B}_1^\mathsf{T}, \hdots, \mathbf{B}_T^\mathsf{T}]^\mathsf{T} 
    &\in \mathbb{R}^{3N\times K}
    \\
    \mathbf{H} &\triangleq\frac{1}{2}\left(\mathbf{B}^\mathsf{T}\left(\sum_{t=1}^T \mathbf{W}_t\right)\mathbf{B} + \lambda \mathbf{I}_K\right)^{-1}
    &\in \mathbb{R}^{K\times K}
    \end{array}
    \end{equation}
    \begin{equation}
        \mathbf{G}\triangleq  \mathbf{H} - \frac{\mathbf{H}\mathbf{1}_K\mathbf{1}_K^\mathsf{T}\mathbf{H}}{\mathbf{1}_K^\mathsf{T}\mathbf{H}\mathbf{1}_K},\qquad\:\mathbf{g}\triangleq \frac{\mathbf{H}\mathbf{1}_K}{\mathbf{1}_K^\mathsf{T} \mathbf{H} \mathbf{1}_K}
    \end{equation}
    \begin{proof}
        Similar to~\cite{Shi23tro-PACE}, see~\prettyref{appendix:shape}.
    \end{proof}
\end{proposition} 

\subsection{Change of Variables to Quadratic Program}
\label{sec:qcqp}
Problem~\eqref{eq:optprob_original} remains non-convex in the state variables $(\BR_t, \Bp_t, \Bv_t, \BOmega_t)$ due to quadratic equality constraints. We aim to relax this problem into a convex semidefinite program. Towards this goal, we show how~\eqref{eq:optprob_original} can be rewritten as a quadratically constrained quadratic program (QCQP). 

First, we observe that the constraint set is already quadratic: the dynamics \eqref{eq:prUpdate} are quadratic equalities and $\SO{3}$ constraints on rotations can be written as quadratic equality constraints, see, \eg~\cite{Tron15rssws3D-dualityPGO3D}. In the objective, however, the squared norm of $\BR_t\mathbf{B}_i\mathbf{c^\star}$ is quartic. We use the rotational invariance of the $\ell_2$ norm to reparametrize position as $\mathbf{s}_t\triangleq\mathbf{R}^\mathsf{T}_t\mathbf{p}_t$, turning the problem into a QCQP. The result is summarized below.

\begin{proposition}[QCQP Formulation]
\label{prop:qcqpform}
    Let $\mathbf{c}$ be defined as in~\eqref{eq:optshape}, and note that it is a linear function of $\mathbf{R}_t$ and $\mathbf{s}_t$.
    The shape estimation and tracking problem can be equivalently written as a quadratically constrained quadratic program:

    \begin{equation}
    \label{eq:optprob_qcqp}
    \begin{aligned}[c]
        \min_{
            \substack{
                \BR_t,\BOmega_l\in \SO{3}, \\
                \mathbf{s}_t,\Bv_l\in\mathbb{R}^{3},\\
                t = 1,\hdots,T,\\
                l = 1,\hdots,T-1
                }
        }
        \\[24pt]
        \textrm{s.t.} & 
    \end{aligned}
    \begin{aligned}[c]
        &\sum_{t=1}^T \sum_{i=1}^N w_t^i \left\| \BR_t^\mathsf{T}\mathbf{y}_t^i - \mathbf{B}_i\mathbf{c} - \mathbf{s}_t\right\|^2 
        + \lambda\|\Delta\mathbf{c}\|^2
        &&\\
        &+ \sum_{t=1}^{T-1} \omega_t\|\Bv_{t+1}-\Bv_t\|^2
        + \kappa_t\|\BOmega_{t+1}-\BOmega_t\|^2_F &&\\
        &\BOmega_t\mathbf{s}_{t+1} = \mathbf{s}_t + \mathbf{v}_t,\:\:
        \mathbf{R}_{t+1} = \mathbf{R}_t\boldsymbol{\Omega}_t&&
    \end{aligned}
    \end{equation}
    \begin{proof}
        See \prettyref{appendix:quartic2qcqp}.
    \end{proof}
\end{proposition}

We rewrite~\eqref{eq:optprob_qcqp} in canonical form, separating the quadratically constrained variables ($\mathbf{s},\BR,\BOmega$) from the linearly constrained ones ($\Bv$):
\begin{equation}
\hspace{-5mm}
    \label{eq:qcqp}
    \begin{aligned}
        f^\star = \minproblem{\substack{
        \mathbf{x}\in\mathbb{R}^{21T-8}\\
        \Bv\in\mathbb{R}^{3T-3}}}
        {\mathbf{x}^\mathsf{T} \mathbf{Q} \mathbf{x} + \Bv^\mathsf{T}\mathbf{P}\Bv}
        {\mathbf{x}^\mathsf{T} \mathbf{A}_i \mathbf{x} + \mathbf{d}_i^\mathsf{T}\mathbf{v} + f_i= 0,\:i=1,\hdots,m}
    \end{aligned}
\end{equation}
In this equation, $\mathbf{x}$ is a vector in homogeneous form stacking all the unknowns in~\eqref{eq:optprob_qcqp} except for $\mathbf{v}_t$ terms, which are stacked in $\mathbf{v}$; $\mathbf{Q}$, $\mathbf{P}$, and $\mathbf{A}_i$ are known symmetric matrices governing the quadratic objective and constraints, and the vectors $\mathbf{d}_i$ and scalars $f_i$ capture the linear and constant portions of the $m$ constraints, respectively.

\subsection{Convex Semidefinite Relaxation}
\label{sec:semidefin}
While the QCQP in~\eqref{eq:qcqp} is still non-convex in the variable $\vxx$, it admits a standard semidefinite relaxation \cite{Briales16iros,Eriksson18cvpr-strongDuality,Rosen18ijrr-sesync}. Instead of solving for $\vxx$ directly, we reparametrize the problem using $\mathbf{X}=\mathbf{xx}^\mathsf{T}$ (a rank-1 positive semidefinite matrix), and drop the rank-1 constraint on $\mathbf{X}$ to obtain a convex problem that may be solved by off-the-shelf solvers such as MOSEK~\cite{mosek}. 
This is the well-known  Shor's relaxation~\cite{Shor87}.

\begin{corollary}[Shor's Relaxation]
    The following semidefinite program (SDP) is a convex relaxation of \eqref{eq:qcqp}:
    \begin{equation}
    \begin{aligned}
        \label{eq:sdp}
        f^\star_{\mathrm{SDP}} = 
        \min_{\substack{
        \mathbf{X}\in\mathbb{S}^{21T-8}\\
        \Bv\in\mathbb{R}^{3T-3}}} 
        & \mathrm{trace}(\mathbf{Q} \mathbf{X}) + \Bv^\mathsf{T}\mathbf{P}\Bv\\
        \text{s.t.} \;\;\;& \mathrm{trace}(A_i \mathbf{X}) + \mathbf{d}_i^\mathsf{T}\mathbf{v} + f_i = 0,\\
        &\mathbf{X} \succeq 0,\;\;i=1,\hdots,m
    \end{aligned}
    \end{equation}
    Further, when the solution $\mathbf{X^\star}$ of~\eqref{eq:sdp} is rank-1 we can recover exactly the solution to the non-convex QCQP~\eqref{eq:qcqp} by factorizing $\mathbf{X^\star}=\mathbf{x^\star(x^\star)^\mathsf{T}}$. 
\end{corollary}

Similar to relaxations derived in related work~\cite{Rosen18ijrr-sesync,Shi21rss-pace,Shi23tro-PACE} the rank of $\mathbf{X^\star}$ is a \emph{certificate} for the optimality of the solution. Moreover, we can bound the suboptimality of a feasible solution to~\eqref{eq:qcqp} obtained from~\eqref{eq:sdp} using the objective. Given a feasible solution $(\hat{\mathbf{x}}, \hat{\mathbf{v}})$ achieving objective $\hat{f}$ in~\eqref{eq:qcqp}, we bound its suboptimality using $\hat{f} - f^\star_{\mathrm{SDP}} \geq \hat{f}-f^\star \geq 0$. The condition $\hat{f} = f^\star_{\mathrm{SDP}}$ also certifies the optimality of the solution. The scalar $\hat{f} - f^\star_{\textrm{SDP}}$ is called the \emph{suboptimality gap}.

%
The relaxation~\eqref{eq:sdp} is relevant in practice because we observe it to be empirically tight in the 
case of low-to-moderate noise and no outliers; hence it can produce optimal solutions without needing an initial guess.
 Moreover, the size of the SDP~\eqref{eq:sdp} is independent of the size of the shape library, hence the relaxation is relatively efficient to solve.
We name the resulting approach \name: \emph{Certifiable Algorithm for Shape estimation and Tracking}. %

\section{Adding Outlier Robustness}
\label{sec:outlier}

Real-world measurements are often corrupted by outliers. In particular, sparse keypoints are vulnerable to misdetections and incorrect depth measurements. Without modifications, outliers degrade the result of \name. %
To tackle this problem, we propose a preprocessing step in which we quickly identify and prune gross outliers, and a wrapper for \name that iteratively converges to the inlier set. We name this approach \nameSharp and show empirical robustness to 50-60\% outliers.

\subsection{Compatibility Checks to Remove Gross Outliers}
\begin{figure}[tb]
\centerline{\includegraphics[width=\linewidth]{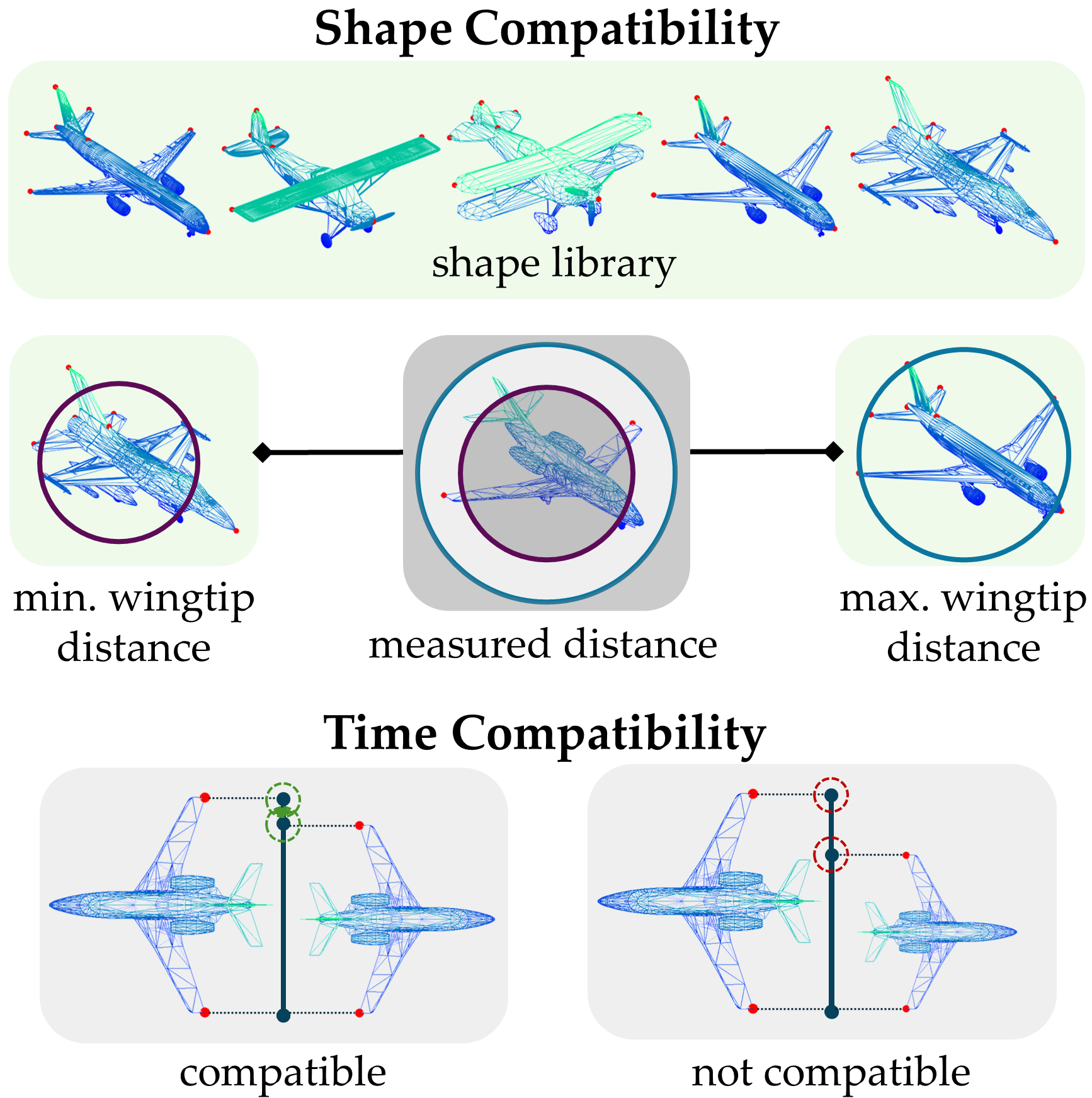}}
\caption{\textbf{Outlier Pruning.} Most outliers are easy to identify via shape or time compatibility tests. Shape compatibility retains keypoints that are mutually within the convex hull of the known shape library. Time compatibility compares keypoint pairs over multiple observations and retains groups that preserve 3D distance over time, up to a tolerance $\epsilon$. We determine the largest set of compatible measurements via a mixed integer linear program.}
\label{fig:pruning}
\end{figure}

Inspired by~\cite{Shi23tro-PACE}, we introduce compatibility tests to identify gross outliers (\prettyref{fig:pruning}).
 These tests rely on the assumptions of rigid-body motion of the object and the active shape model.
The most likely inlier set is thus the largest set of compatible measurements, found via a fast mixed-integer linear program. 

\textbf{Shape Compatibility.} Recall that any observed object must lie within the convex hull of the shape library by assumption. Framed as pairwise compatibility, the true distance between any two keypoints $i$ and $j$ must lie somewhere between the minimum and maximum distance between $i$ and $j$ in the shape library models. Therefore, any two keypoint measurements that are outside this bound cannot simultaneously be inliers; one or both must be outliers. Allowing for keypoint noise expands these bounds as summarized in \prettyref{prop:shapecompatibility}. Refer to \cite{Shi21rss-pace} for a full proof.

\begin{proposition}[Shape Compatibility Test]
    \label{prop:shapecompatibility}
    Let $\epsilon$ be the maximum error for a measurement to be considered an inlier. If a pair of measurements $\mathbf{y}_t^i$ and $\mathbf{y}_t^j$ are both inliers, then:
    \begin{equation}
        \label{eq:shapecomp}
        \begin{aligned}
            b_{ij}^{\mathrm{min}}-2\epsilon 
            \leq
            \|\mathbf{y}_t^i - \mathbf{y}_t^j\|
            \leq
            b_{ij}^{\mathrm{max}}+2\epsilon
        \end{aligned}
    \end{equation}
    where $b_{ij}^{\{\mathrm{min},\mathrm{max}\}}$ are the minimum and maximum distances between keypoints $i$ and $j$ in the shape library:
    \begin{equation}
        b_{ij}^{\{\mathrm{min},\mathrm{max}\}}\triangleq
        \{\underset{\mathbf{c}\geq 0, \mathbf{1}^T\mathbf{c}=1}{\min,\max}\}\|(\mathbf{B}_i - \mathbf{B}_j)\mathbf{c}\|
    \end{equation}
\end{proposition}

\textbf{Time Compatibility.} For a rigid body the distance between two points is constant over time. This forms the basis for a compatibility test between pairs of points at two times.

\begin{proposition}[Time Compatibility Test]
    \label{prop:timecompatibility}
    Let $\epsilon$ be the maximum error for a measurement to be considered an inlier. Consider the measurements of keypoints $i$ and $j$ at times $l$ and $m$. If these measurements are all inliers then:
    \begin{equation}
        \label{eq:timecomp}
        \begin{aligned}
            |\|\mathbf{y}_l^i - \mathbf{y}_l^j\| - \|\mathbf{y}_m^i - \mathbf{y}_m^j\||
            \leq
            4\epsilon
        \end{aligned}
    \end{equation}
    See \prettyref{appendix:timecompproof} for proof.
\end{proposition}

\textbf{Outlier Pruning.} Any set of inliers must satisfy the compatibility conditions presented above. To prune gross outliers, we select the largest set of compatible measurements. Finding this set can be cast as a mixed-integer linear program which we solve using COPT~\cite{copt}.

\begin{proposition}[Largest Set of Compatible Measurements]
    \label{prop:milp}
    Let $\mathcal{S}$ be the set of measurement pairs that do not satisfy the shape compatibility condition~\eqref{eq:shapecomp} and $\mathcal{T}$ be the set of groups of four measurements that do not satisfy the time compatibility condition~\eqref{eq:timecomp}.
    The largest set of measurements that satisfy both shape and time compatibility is given by the following mixed integer linear program:
    \begin{equation}
    \begin{aligned}
        \mathop{\arg\!\max}\limits_{\boldsymbol{\theta}\in\{0,1\}^{N\times T}}
        \: &
        \sum_{t=1}^T \sum_{i=1}^{N} \theta_t^i &\\
        \textrm{s.t.}\quad & \theta_t^i + \theta_t^j \leq 1 \quad \forall\ (t,i,j)\in\mathcal{S} \\
        & \theta_l^i + \theta_l^j + \theta_m^i + \theta_m^j \leq 3 \quad \forall\ (l,m,i,j)\in\mathcal{T}
    \end{aligned}
    \end{equation}
    where $\theta_t^i = 1$ denotes including measurement $\mathbf{y}_t^i$ in the set.
\end{proposition}
The proof follows from Propositions~\ref{prop:shapecompatibility} and~\ref{prop:timecompatibility}.

\subsection{Graduated Non-Convexity for Robustness.}
While consistency checks can remove a significant proportion of outliers, they may miss a number of difficult-to-detect outliers. To remove these remaining outliers we use \name as a non-minimal solver for \textit{graduated non-convexity} (GNC)~\cite{Yang20ral-GNC}. 
We use the truncated least squares loss in GNC and follow the implementation and parameter choices of~\cite{Yang20ral-GNC}.
 In our experiments, we show the combination of GNC and our compatibility checks is robust to 50-60\% of outliers. 

\begin{figure*}[tb]
    \centering
    \subfloat[Measurement noise robustness of \name with fixed process noise.]{\includegraphics[width=\linewidth]{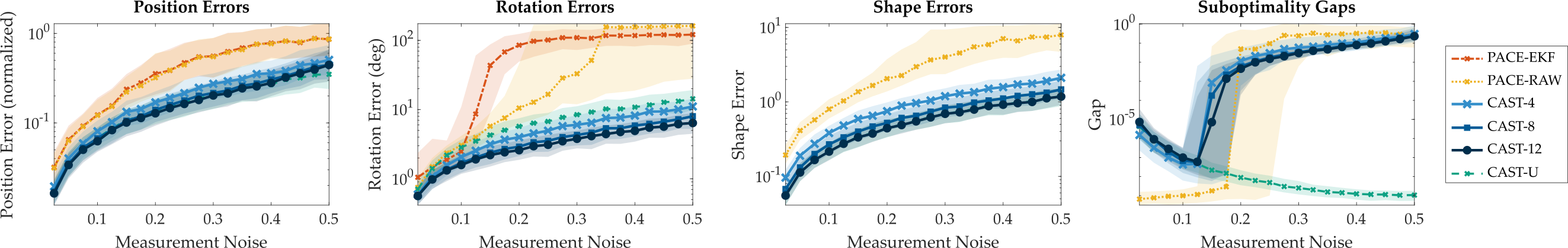}} \\
    \subfloat[Process noise robustness of \name with fixed measurement noise.]{\includegraphics[width=\linewidth]{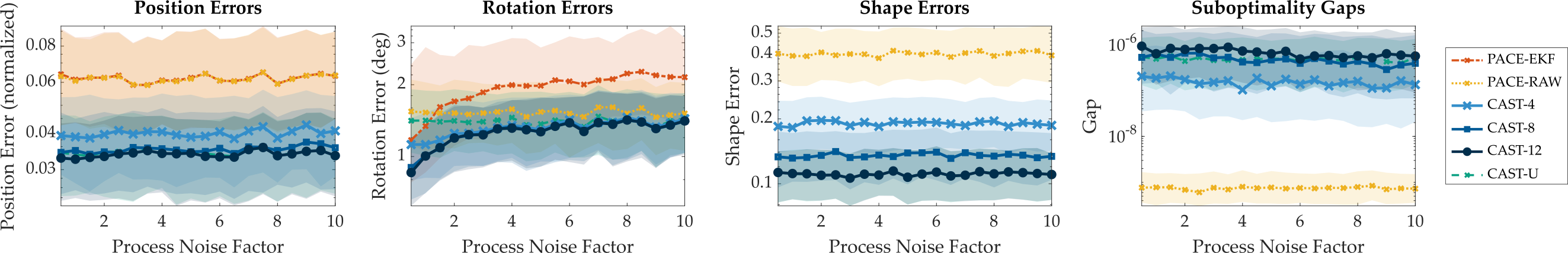}} \\
    \subfloat[Outlier robustness of \nameSharp and ablations.]{\includegraphics[width=\linewidth]{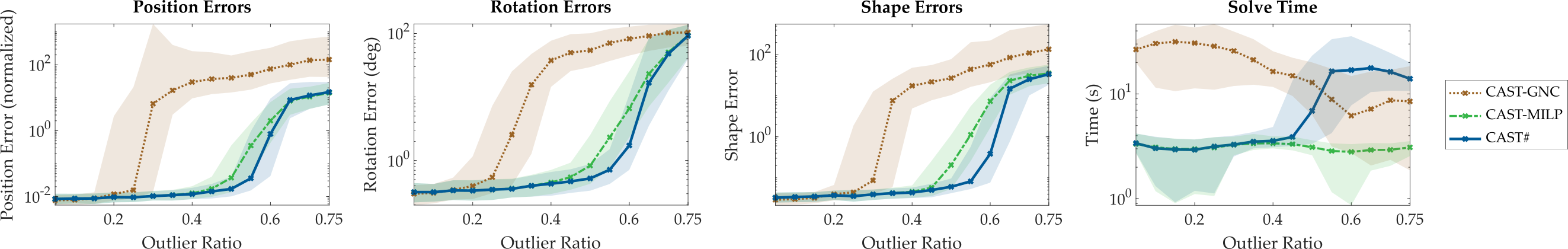}}
    \caption{ \textbf{Performance of \name and \nameSharp in Synthetic Experiments.} Using the PASCAL3D+ aeroplane shape library, we generate synthetic measurements to test the robustness of \name and \nameSharp to measurement noise, process noise, and outliers. Plots show median and IQR of 500 runs.\vspace{-7pt}}
    \label{fig:synthetic_results}
    \vspace*{-5pt}
\end{figure*}
\vspace*{-5pt}
\section{Experiments}
\label{sec:experiments}
This section characterizes CAST. Synthetic experiments (Section~\ref{sec:exp:synthetic}) show the semidefinite relaxation in \name is empirically tight and returns accurate estimates in the presence of noise, while \nameSharp is robust to 50-60\%  outliers. Sections~\ref{sec:exp:ycbineoat}, \ref{sec:exp:nocs}, and \ref{sec:exp:drone} show \nameSharp\xspace{is competitive with other category-level approaches on two public datasets and a real-world drone-based vehicle tracking scenario.}


\subsection{Optimality and Robustness in Synthetic Dataset}
\label{sec:exp:synthetic}

\textbf{Dataset.} We generate keypoint measurements according to the measurements and motion models in \prettyref{sec:problem}. Ground truth trajectories follow the constant twist model~\eqref{eq:velUpdate} with Gaussian velocity noise and Langevin rotation rate noise (process noise). The ground truth trajectory and randomly generated shape determine the measured keypoint positions without regard for occlusion, subject to Gaussian perturbations (measurement noise) and outliers. We use the realistic PASCAL3D+ aeroplane shape library~\cite{Xiang14-Pascal3DPlus} (with characteristic length $l=0.2$ m) to generate a ground truth shape vector. {In each experiment, we fix measurement noise to $5$\% of the characteristic length, and the process noise to $0.01$ m and {$0.01$ rad}. For the measurement noise experiment, we set the velocity weights $\omega_t = 1$ to improve tightness;} results with standard weights are provided in \prettyref{appendix:exp_results}.

\textbf{Baselines.} We compare \name against PACE~\cite{Shi23tro-PACE}, a certifiably optimal solver for single-frame pose estimation, and PACE+EKF, an approach that filters the pose estimate from PACE at each time using an extended Kalman filter (EKF) while using a constant-twist motion model. We test \name with a time horizon of $4$, $8$, or $12$ frames and label the corresponding results as
\nameRaw-4, \nameRaw-8, and \nameRaw-12; we also report \nameRaw-U, which is a variant of \nameRaw-12 with no velocity or rotation smoothing ($\omega_t = 0$, $\kappa_t$ = 0). 
Finally, to test \nameSharp, we replace a fraction of the measurements with random outliers normally distributed about the centroid of the object with a standard deviation equal to the characteristic length. 
For the tests with outliers, we compare against ablations of \nameSharp with only GNC or only compatibility-based (MILP) outlier rejection without GNC and use $T=12$. 

\textbf{Results.} Fig~\ref{fig:synthetic_results} reports the {median} position error (as a percent of length scale), rotation error (in degrees), shape error ($l_2$ distance between predicted and ground truth shape vector $\mathbf{c}$), and suboptimality gap or solve time for increasing measurement noise (normalized by length scale), process noise (reported as a multiple of $5$\%), and outlier ratio. In all experiments, 
\name and \nameSharp achieve the lowest median position, rotation, and shape error. In Figs.~\ref{fig:synthetic_results}(a) and~\ref{fig:synthetic_results}(b), \name is consistently tight (suboptimality gap $< 10^{-4}$) in low to moderate noise, and still gives an accurate estimate when not tight. Interestingly, while \name outperforms its unsmoothed variant \nameRaw-U, the latter remains tight for higher measurement noise. \nameRaw-U benefits over PACE from access to additional measurements, and \name benefits over \nameRaw-U from additional information about the object's motion. The primary cost of \name compared to PACE is its runtime, which ranged from 0.1 to 7 Hz depending on the time horizon; see \prettyref{appendix:exp_results} for detailed runtimes.

We also note the poor performance of PACE-EKF in both experiments. The EKF provides some benefit for very low noise but quickly diverges for higher noise as the distribution of PACE measurements deviates from Gaussian and the dynamics are nonlinear. We provide a comparison to an EKF using perturbed ground truth poses in~\prettyref{appendix:exp_results}.%

The outlier experiment in \prettyref{fig:synthetic_results}(c) shows robustness to 50-60\% of outliers using \nameSharp. Compatibility tests alone are robust to 40-50\% of outliers, while GNC only tolerates 20-30\% of outliers. The data show GNC and MILP-based outlier rejection are complimentary, with the fast MILP solve time being unaffected by GNC in the low outlier regime.

\subsection{YCBInEOAT Dataset}
\label{sec:exp:ycbineoat}

\begin{table}[t]
    \centering
    \caption{Comparison of Methods on YCBInEOAT Dataset\vspace{-5pt}}
    \label{tab:ycbineoat}
    \resizebox{\columnwidth}{!}{\begin{tabular}{c|cc|cc|cc||c}
    \hline
    \multirow{2}{*}{\textbf{Method}} & \multicolumn{2}{c|}{Cracker}                         & \multicolumn{2}{c|}{Sugar}                           & \multicolumn{2}{c||}{Mustard} & \multicolumn{1}{c}{\hspace{-2pt}Reconst.\hspace{-2pt}}\\ 
                            & \multicolumn{1}{l}{\hspace{-2pt}ADD} & \multicolumn{1}{l|}{\hspace{-8pt}ADD-S\hspace{-5pt}} & \multicolumn{1}{l}{\hspace{-2pt}ADD} & \multicolumn{1}{l|}{\hspace{-8pt}ADD-S\hspace{-5pt}} & \multicolumn{1}{l}{\hspace{-2pt}ADD} & \multicolumn{1}{l||}{\hspace{-8pt}ADD-S\hspace{-5pt}} & \multicolumn{1}{l}{\hspace{-2pt}CD\:(cm)\hspace{-2pt}}\\ \hline& & & & & &\\[-1.8ex]
    6-PACK                  & \hspace{-2pt}-                       & \hspace{-8pt}-\hspace{-5pt}                          & \hspace{-2pt}-                       & \hspace{-8pt}-\hspace{-2pt}                          & \hspace{-2pt}34.49                   & \hspace{-8pt}80.76\hspace{-5pt} & \hspace{-2pt}-\\
    TEASER++                & \hspace{-2pt}84.76                   & \hspace{-8pt}\underline{92.14}\hspace{-5pt}                      & \hspace{-2pt}83.26                  & \hspace{-8pt}\underline{91.27}\hspace{-5pt}                     & \hspace{-2pt}86.02                   & \hspace{-8pt}93.43\hspace{-5pt} & \hspace{-2pt}-\hspace{-2pt}\\
    MaskFusion              & \hspace{-2pt}79.74                   & \hspace{-8pt}88.28\hspace{-5pt}                      & \hspace{-2pt}36.18                   & \hspace{-8pt}45.62\hspace{-5pt}                      & \hspace{-2pt}11.55                   & \hspace{-8pt}13.11\hspace{-5pt} & \hspace{-2pt}-\hspace{-2pt}\\
    BundleTrack             & \hspace{-2pt}\underline{85.07}          & \hspace{-8pt}89.41\hspace{-5pt}                      & \hspace{-2pt}\underline{85.56}                   & \hspace{-8pt}{90.22}\hspace{-5pt}                      & \hspace{-2pt}\textbf{92.26}          & \hspace{-8pt}\underline{95.35}\hspace{-5pt} & \hspace{-2pt}2.81\hspace{-2pt}\\
    BundleSDF              & \hspace{-2pt}\hspace{-2pt}81.44                   & \hspace{-8pt}90.76\hspace{-5pt}                      & \hspace{-2pt}\textbf{86.55}          & \hspace{-8pt}\textbf{92.85}\hspace{-5pt}             & \hspace{-2pt}\underline{90.83}                  & \hspace{-8pt}\textbf{95.48}\hspace{-5pt} & \hspace{-2pt}1.16\hspace{-2pt}\\
    \nameSharp-8                    & \hspace{-2pt}\textbf{86.93}                   & \hspace{-8pt}\textbf{93.14}\hspace{-5pt}             & \hspace{-2pt}81.97                  & \hspace{-8pt}89.45\hspace{-5pt}                     & \hspace{-2pt}84.67                   & \hspace{-8pt}92.41\hspace{-5pt} & \hspace{-2pt}\textbf{0.16}\hspace{-2pt}\\ \hhline{=|==|==|==||}
    & & & & & &\\[-1.8ex]
    \nameSharp-GTK           & \hspace{-2pt}89.00                   & \hspace{-8pt}94.09\hspace{-5pt}             & \hspace{-2pt}91.05                  & \hspace{-8pt}95.27\hspace{-5pt}                     & \hspace{-2pt}92.18                   & \hspace{-8pt}96.17\hspace{-5pt} & \hspace{-2pt}-\hspace{-2pt}\\ \hline
    \end{tabular}}
\end{table}

\textbf{Dataset.} The YCBInEOAT dataset~\cite{Wen20iros-seTrack} includes 9 RGB-D videos of a robotic manipulator interacting with 5 YCB objects~\cite{Calli15-YCBobject}. 
We train a simple RGB keypoint detector for each object using their CAD models and manually-defined semantic points. The detector has a ResNet18 backbone~\cite{He16cvpr-ResNet} and is trained on the BOP YCB-V synthetic dataset~\cite{Denninger23joss-blenderproc}. We report the ADD and ADD-S area under the precision-accuracy curve (AUC) scores at $0.1$ m with estimated poses applied to ground truth CAD models; see~\cite{Xiang17rss-posecnn}.

\textbf{Baselines.}
We compare against state-of-the-art instance and category-level tracking approaches for the cracker, sugar, and mustard objects. We omit the small soup object and bleach because it matches the background and gripper colors and our simple keypoint detector is unable to detect reasonable keypoints. TEASER++~\cite{Yang20tro-teaser} is the only instance-level approach and uses the same keypoints given to \nameSharp.
{
    For \nameSharp, we group mustard and bleach into the ``bottle'' category along with a CAD model of a ketchup bottle~\cite{Smit20-ketchup} (3 shapes, 65 keypoints). Similarly, we group cracker and sugar into the ``box'' category (2 shapes, 52 keypoints).
}
Results from 6-PACK~\cite{Wang20icra-6pack}, MaskFusion~\cite{Runz18ismar-maskfusion}, and BundleTrack~\cite{Wen21iros-bundletrack} are taken from the results reported in \cite{Wen21iros-bundletrack}. BundleSDF~\cite{Wen23cvpr-bundlesdf} results were replicated using the open source implementation with ground truth segmentation masks.
Since our keypoint detector is fairly simple, 
we also report \nameSharp evaluated on ground truth pixel keypoints with true depth and occlusions.
For all methods we compute scores using the ground truth shape and initialize with the first frame ground truth pose. For \nameSharp and TEASER we do not use any initialization. We also report the chamfer distance between the tightest final shape estimate (a dense 3D model) and the true shapes averaged over all 5 objects~\cite{Wen23cvpr-bundlesdf}.

\textbf{Results.} {\nameSharp clearly outperforms baselines for the cracker object but underperforms for sugar and mustard (\prettyref{tab:ycbineoat}). The results are encouraging: even with a simple keypoint detector, \nameSharp outperforms elaborate learning-based methods. The sugar and mustard results are not far behind baselines and mostly reflect the quality of the keypoint detector, which struggles with smaller objects (see \href{https://youtu.be/eTIlVD9pDtc}{video attachment}). Given ground truth pixel keypoints, \nameSharp outperforms virtually all baselines, despite the low-quality depth data in the dataset. CAST\# achieves a near-perfect reconstruction chamfer distance; note that the shape library includes the true model.\vspace{-2pt}}

\subsection{NOCS Dataset}
\label{sec:exp:nocs}
\textbf{Dataset and Baselines.} We evaluate our system on NOCS-REAL275~\cite{Wang19-normalizedCoordinate} on the camera (2561 frames) and mug (2615 frames) object categories. We use a keypoint detector based on YOLOv8-pose~\cite{yolov8} trained on synthetic images of 5 mugs and 3 cameras compiled from 3D scans because the NOCS training data does not include precise ground truth. Different from YCBInEOAT, the detector and shape library do not include the ground truth object models. We drop the laptop due to bad training data and the symmetric objects, which CAST is not designed to handle. We report $\mathbf{5^{\boldsymbol{\circ}}5}$\textbf{cm}: the percent of estimates within $5^\circ$ and $5$ cm of the ground truth, $\mathrm{\mathbf{R_{err}}}$: the mean orientation error (degrees), and $\mathrm{\mathbf{p_{err}}}$: the mean position error (cm). Consistent with~\cite{Wang20icra-6pack}, we exclude measurements with high position error ($0.1$\:m) from $\mathrm{{R_{err}}}$ and $\mathrm{{p_{err}}}$. We compare category-level pose tracking with 6-PACK~\cite{Wang20icra-6pack}, CAPTRA~\cite{Weng21iccv-captra}, BundleTrack~\cite{Wen21iros-bundletrack}, and iCaps~\cite{Deng22ral-icaps}. iCaps and \nameSharp consider the more difficult problems of initializing tracking with a segmentation mask or 2D detection, respectively. 
Baselines are from Table\:1 in~\cite{Deng22ral-icaps}. 

\textbf{Results.} \nameSharp achieves state of the art
among methods that do not initialize via ground truth (\prettyref{tab:nocs}). There is still substantial gap between \nameSharp and methods that rely on the less practical assumption of ground truth initialization. This reflects the quality of the keypoint detector; with ground truth 3D keypoints ignoring occlusions, \nameSharp is nearly perfect and outperforms all baselines except in $p_\mathrm{err}$, which is due to the small shape library.

\begin{table}[]
    \centering
    \caption{Comparison of Methods on NOCS Dataset\vspace{-5pt}}
    \label{tab:nocs}
    \resizebox{\columnwidth}{!}{\begin{tabular}{lc|ccc||cc||c}
    \hline \multicolumn{2}{l|}{}&&&\multicolumn{1}{c||}{}\\[-2.ex]
    \multicolumn{2}{l|}{\textbf{Method}}                                                & \hspace{-4pt}6Pack   & \hspace{-7pt}CAPTRA  & \hspace{-8pt}BundleTrack\hspace{-4pt} & \hspace{-3pt}iCaps           & \hspace{-8pt}{CAST\#-8}\hspace{-3pt} & \hspace{-4pt}{CAST\#-GT}\hspace{-3pt}\\ \hline \multicolumn{2}{l|}{}&&&\multicolumn{1}{c||}{}&&\multicolumn{1}{c||}{}\\[-2.ex]
    \multicolumn{2}{l|}{Initialization}                                        & \hspace{-4pt}Pert GT & \hspace{-7pt}Pert GT & \hspace{-8pt}Pert GT\hspace{-4pt}     & \hspace{-3pt}2D seg. & \hspace{-8pt}\textbf{2D det.}\hspace{-3pt} & \hspace{-4pt}{2D det.}\hspace{-3pt}\\ \hline \multicolumn{1}{c|}{}&\multicolumn{1}{c|}{}&&&\multicolumn{1}{c||}{}&&\multicolumn{1}{c||}{}\\[-2.ex]
    \multicolumn{1}{c|}{\multirow{3}{*}{\rotatebox[origin=c]{90}{camera}}} & \hspace{-5pt}$5^{\circ}5$cm\hspace{-5pt}  & \hspace{-4pt}10.1    & \hspace{-7pt}0.41    & \hspace{-8pt}\textbf{85.8}\hspace{-4pt}        & \hspace{-3pt}9.32            & \hspace{-8pt}\underline{20.68}\hspace{-3pt}  & \hspace{-4pt}{88.42}\hspace{-3pt}           \\
    \multicolumn{1}{c|}{}                        & \hspace{-5pt}$\mathrm{R}_{\mathrm{err}}$\:(deg)\hspace{-5pt} & \hspace{-4pt}35.7    & \hspace{-7pt}17.82   & \hspace{-8pt}\textbf{3.0}\hspace{-4pt}           & \hspace{-3pt}13.69           & \hspace{-8pt}\underline{11.07}\hspace{-3pt} & \hspace{-4pt}{2.33}\hspace{-3pt}            \\
    \multicolumn{1}{c|}{}                        & \hspace{-5pt}$\mathrm{p}_{\mathrm{err}}$\:(cm)\hspace{-5pt} & \hspace{-4pt}5.6     & \hspace{-7pt}35.53   & \hspace{-8pt}\textbf{2.1}\hspace{-4pt}         & \hspace{-3pt}\underline{2.72}            & \hspace{-8pt}3.81\hspace{-3pt}   &  \hspace{-4pt}{2.55}\hspace{-3pt}        \\ \hline \multicolumn{1}{c|}{}&\multicolumn{1}{c|}{}&&&\multicolumn{1}{c||}{}&&\multicolumn{1}{c||}{}\\[-2.ex]
    \multicolumn{1}{l|}{\multirow{3}{*}{\rotatebox[origin=c]{90}{mug}}}    & \hspace{-5pt}$5^{\circ}5$cm\hspace{-5pt}      & \hspace{-4pt}24.1    & \hspace{-7pt}\textbf{55.17}   & \hspace{-8pt}\underline{53.6}\hspace{-4pt}        & \hspace{-3pt}21.82           & \hspace{-8pt}35.22\hspace{-3pt}      & \hspace{-4pt}{97.71}\hspace{-3pt}       \\
    \multicolumn{1}{l|}{}                        & \hspace{-5pt}$\mathrm{R}_{\mathrm{err}}$\:(deg)\hspace{-5pt} & \hspace{-4pt}21.3    & \hspace{-7pt}\underline{5.36}    & \hspace{-8pt}\textbf{5.2}\hspace{-4pt}         & \hspace{-3pt}10.69           & \hspace{-8pt}8.42\hspace{-3pt}    & \hspace{-4pt}{1.45}\hspace{-3pt}          \\
    \multicolumn{1}{l|}{}                        & \hspace{-5pt}$\mathrm{p}_{\mathrm{err}}$\:(cm)\hspace{-5pt} & \hspace{-4pt}2.3     & \hspace{-7pt}\textbf{0.79}    & \hspace{-8pt}2.2\hspace{-4pt}         & \hspace{-3pt}\underline{1.31}            & \hspace{-8pt}2.46\hspace{-3pt}     & \hspace{-4pt}{2.18}\hspace{-3pt}        \\ \hline
    \end{tabular}}
\end{table}

\begin{table}[tbp]
    \vspace{-5pt}
    \centering
    \caption{Quantitative Results of Drone Experiment\vspace{-5pt}}
    \label{tab:drone}
    \begin{tabular}{c|ccccc}
    \hline\\[-2.ex]
    \textbf{Method} & \hspace{-1pt}ADD $\uparrow$ & \hspace{-5pt}$R_{\mathrm{err}}$ (deg) $\downarrow$ & \hspace{-3pt}$p_{\mathrm{err}}$ (cm) $\downarrow$ & \hspace{-5pt}$c_\mathrm{err}$ $\downarrow$ & \hspace{-5pt}FPS $\uparrow$                 \\ \hline& \\[-1.8ex]
    TEASER~\cite{Yang20tro-teaser}          & \hspace{-1pt}57.0             & \hspace{-5pt}$9.6\pm23.2$                          & \hspace{-3pt}$4.3\pm3.8$                          & \hspace{-5pt}-                             &   \hspace{-5pt}\textbf{39.1}                           \\
    PACE~\cite{Shi21rss-pace}            & \hspace{-1pt}52.0             & \hspace{-5pt}$12.1\pm32.0$                         & \hspace{-3pt}$3.2\pm2.4$                          & \hspace{-5pt}0.79                          & \hspace{-5pt}3.94 \\
    CAST\#-4          & \hspace{-1pt}56.6             & \hspace{-5pt}$7.6\pm4.5$                           & \hspace{-3pt}$\mathbf{2.7}\pm1.3$                          & \hspace{-5pt}0.84                          & \hspace{-5pt}3.65                           \\
    CAST\#-8          & \hspace{-1pt}58.0             & \hspace{-5pt}$7.0\pm4.3$                           & \hspace{-3pt}$\mathbf{2.7}\pm1.4$                          & \hspace{-5pt}0.76                          & \hspace{-5pt}1.44                           \\
    CAST\#-12         & \hspace{-1pt}\textbf{58.8}             & \hspace{-5pt}$\mathbf{6.5}\pm 3.8$                          & \hspace{-3pt}$\mathbf{2.7}\pm1.4$                          & \hspace{-5pt}\textbf{0.71}                          & \hspace{-5pt}0.67                           \\
    CAST\#-U          & \hspace{-1pt}58.2             & \hspace{-5pt}$8.8\pm 15.1$                         & \hspace{-3pt}$4.6\pm20.0$                           & \hspace{-5pt}\textbf{0.71}                          & \hspace{-5pt}0.91                           \\ \hline
    \end{tabular}
\end{table}

\subsection{Drone-based Vehicle Tracking}
\label{sec:exp:drone}

We use the drone platform described in~\cite{Ubellacker24npj-softDrone3} to evaluate \nameSharp under dynamic real-world conditions, see the \href{https://youtu.be/eTIlVD9pDtc}{video attachment}.
During the experiment we remotely piloted a mini racecar in an elliptical trajectory while the Soft Drone autonomously followed using the centroid and heading derived from raw keypoints estimated at $30$ Hz. Our keypoint detector, like the YCBInEOAT experiments, used a ResNet-based architecture~\cite{He16cvpr-ResNet} with 7 keypoints, and was trained on images of a similar racecar. Offline, we used motion capture to transform the 3D position of each keypoint to the world frame {(to compensate for the known motion of the drone)} and ran \nameSharp to estimate the racecar's shape and pose at each time step. For the category-level shape library we used scaled PASCAL3D+ car shapes and the racecar instance.

Quantitative results of TEASER, PACE, \nameSharp, and variants are given in \prettyref{tab:drone}. TEASER and PACE operate on the same raw keypoint data as \nameSharp and are tuned for optimal performance. Across metrics, \nameSharp achieves the highest accuracy and lowest mean errors. In particular, the batch approach with motion priors significantly decreases the standard deviation of rotation and position errors. While the frames per second is not competitive with TEASER, \nameRaw-4 is not significantly slower than PACE despite \nameSharp using an unoptimized MATLAB implementation. 
\section{Conclusion}
We propose \name, a certifiably optimal approach to simultaneously estimate the shape and track the pose of an object of a given category. Our approach uses a semidefinite relaxation to solve the tracking problem, avoiding assumptions of small motion and local minima from local solvers. 
In the presence of outliers, \nameSharp uses GNC and a fast shape and time compatibility scheme to reject gross measurements. Our experiments show the relaxation is empirically tight and is competitive against baselines on the YCBInEOAT dataset and in a real-world drone tracking setting.

\scriptsize{
\bibliographystyle{IEEEtranN}

\begin{thebibliography}{55}
\providecommand{\natexlab}[1]{#1}
\providecommand{\url}[1]{#1}
\csname url@samestyle\endcsname
\providecommand{\newblock}{\relax}
\providecommand{\bibinfo}[2]{#2}
\providecommand{\BIBentrySTDinterwordspacing}{\spaceskip=0pt\relax}
\providecommand{\BIBentryALTinterwordstretchfactor}{4}
\providecommand{\BIBentryALTinterwordspacing}{\spaceskip=\fontdimen2\font plus
\BIBentryALTinterwordstretchfactor\fontdimen3\font minus \fontdimen4\font\relax}
\providecommand{\BIBforeignlanguage}[2]{{%
\expandafter\ifx\csname l@#1\endcsname\relax
\typeout{** WARNING: IEEEtranN.bst: No hyphenation pattern has been}%
\typeout{** loaded for the language `#1'. Using the pattern for}%
\typeout{** the default language instead.}%
\else
\language=\csname l@#1\endcsname
\fi
#2}}
\providecommand{\BIBdecl}{\relax}
\BIBdecl

\bibitem[Fortmann et~al.(1980)Fortmann, Bar-Shalom, and Scheffe]{Fortmann80}
T.~Fortmann, Y.~Bar-Shalom, and M.~Scheffe, ``Multi-target tracking using joint probabilistic data association,'' in \emph{Proc. 19th IEEE Conf. on Decision \& Control}, 1980.

\bibitem[Peng(2023)]{peng23-trackingDriving}
Y.~Peng, ``Deep learning for 3d object detection and tracking in autonomous driving: A brief survey,'' 2023.

\bibitem[Wen et~al.(2020)Wen, Mitash, Ren, and Bekris]{Wen20iros-seTrack}
B.~Wen, C.~Mitash, B.~Ren, and K.~E. Bekris, ``se(3)-tracknet: Data-driven 6d pose tracking by calibrating image residuals in synthetic domains,'' in \emph{IEEE/RSJ Intl. Conf. on Intelligent Robots and Systems (IROS)}.\hskip 1em plus 0.5em minus 0.4em\relax IEEE, 2020, pp. 10\,367--10\,373.

\bibitem[Lepetit and Fua(2005)]{Lepetit05-monocularTracking}
V.~Lepetit and P.~Fua, \emph{Monocular Model-Based 3D Tracking of Rigid Objects: A Survey}.\hskip 1em plus 0.5em minus 0.4em\relax Now Foundations and Trends, 2005.

\bibitem[Bruns and Jensfelt(2023)]{Bruns24ras-posesurvey}
L.~Bruns and P.~Jensfelt, ``RGB-D-based categorical object pose and shape estimation: Methods, datasets, and evaluation,'' \emph{Robotics and Autonomous Systems}, vol.~168, 2023.

\bibitem[Shi et~al.(2023)Shi, Yang, and Carlone]{Shi23tro-PACE}
J.~Shi, H.~Yang, and L.~Carlone, ``Optimal and robust category-level perception: Object pose and shape estimation from {2D and 3D} semantic keypoints,'' \emph{{IEEE} Trans. Robotics}, vol.~39, no.~5, pp. 4131--4151, 2023, \linkToPdf{https://arxiv.org/pdf/2206.12498.pdf}.
\bibitem[Wang et~al.(2019)Wang, Sridhar, Huang, Valentin, Song, and Guibas]{Wang19-normalizedCoordinate}
H.~Wang, S.~Sridhar, J.~Huang, J.~Valentin, S.~Song, and L.~Guibas, ``Normalized object coordinate space for category-level 6d object pose and size estimation,'' in \emph{IEEE Conf. on Computer Vision and Pattern Recognition (CVPR)}, 2019, pp. 2642--2651.

\bibitem[Yang et~al.(2020{\natexlab{a}})Yang, Shi, and Carlone]{Yang20tro-teaser}
H.~Yang, J.~Shi, and L.~Carlone, ``{TEASER: Fast and Certifiable Point Cloud Registration},'' \emph{{IEEE} Trans. Robotics}, vol.~37, no.~2, pp. 314--333, 2020, extended arXiv version 2001.07715 \linkToPdf{https://arxiv.org/pdf/2001.07715.pdf}.

\bibitem[Wang et~al.(2023)Wang, Yan, Zhen, Liu, Zhang, Zhang, and Zhou]{Wang23iccv-deepac}
L.~Wang, S.~Yan, J.~Zhen, Y.~Liu, M.~Zhang, G.~Zhang, and X.~Zhou, ``Deep active contours for real-time 6-dof object tracking,'' in \emph{Proceedings of the IEEE/CVF International Conference on Computer Vision}, 2023.

\bibitem[Deng et~al.(2019)Deng, Mousavian, Xiang, Xia, Bretl, and Fox]{Deng2019rss-poserbpf}
X.~Deng, A.~Mousavian, Y.~Xiang, F.~Xia, T.~Bretl, and D.~Fox, ``Poserbpf: A rao-blackwellized particle filter for 6d object pose tracking,'' in \emph{Robotics: Science and Systems (RSS)}, 2019.

\bibitem[Weng et~al.(2021)Weng, Wang, Zhou, Qin, Duan, Fan, Chen, Su, and Guibas]{Weng21iccv-captra}
\BIBentryALTinterwordspacing
Y.~Weng, H.~Wang, Q.~Zhou, Y.~Qin, Y.~Duan, Q.~Fan, B.~Chen, H.~Su, and L.~J. Guibas, ``Captra: Category-level pose tracking for rigid and articulated objects from point clouds,'' in \emph{2021 IEEE/CVF International Conference on Computer Vision (ICCV)}.\hskip 1em plus 0.5em minus 0.4em\relax IEEE, Oct. 2021. [Online]. Available: \url{http://dx.doi.org/10.1109/ICCV48922.2021.01296}
\BIBentrySTDinterwordspacing

\bibitem[Sun et~al.(2022)Sun, Wang, Feng, Wang, Zhao, Stachniss, and Chen]{Sun22iros-icktrack}
\BIBentryALTinterwordspacing
J.~Sun, Y.~Wang, M.~Feng, D.~Wang, J.~Zhao, C.~Stachniss, and X.~Chen, ``Ick-track: A category-level 6-dof pose tracker using inter-frame consistent keypoints for aerial manipulation,'' in \emph{2022 IEEE/RSJ International Conference on Intelligent Robots and Systems (IROS)}.\hskip 1em plus 0.5em minus 0.4em\relax IEEE, Oct. 2022. [Online]. Available: \url{http://dx.doi.org/10.1109/IROS47612.2022.9982183}
\BIBentrySTDinterwordspacing

\bibitem[Wang et~al.(2020)Wang, Mart{\'\i}n-Mart{\'\i}n, Xu, Lv, Lu, Fei-Fei, Savarese, and Zhu]{Wang20icra-6pack}
C.~Wang, R.~Mart{\'\i}n-Mart{\'\i}n, D.~Xu, J.~Lv, C.~Lu, L.~Fei-Fei, S.~Savarese, and Y.~Zhu, ``6-pack: Category-level 6d pose tracker with anchor-based keypoints,'' in \emph{IEEE Intl. Conf. on Robotics and Automation (ICRA)}.\hskip 1em plus 0.5em minus 0.4em\relax IEEE, 2020, pp. 10\,059--10\,066.

\bibitem[Wen et~al.(2023)Wen, Tremblay, Blukis, Tyree, Muller, Evans, Fox, Kautz, and Birchfield]{Wen23cvpr-bundlesdf}
B.~Wen, J.~Tremblay, V.~Blukis, S.~Tyree, T.~Muller, A.~Evans, D.~Fox, J.~Kautz, and S.~Birchfield, ``Bundlesdf: Neural 6-dof tracking and 3d reconstruction of unknown objects,'' \emph{CVPR}, 2023.

\bibitem[R{\"u}nz et~al.(2018)R{\"u}nz, Buffier, and Agapito]{Runz18ismar-maskfusion}
M.~R{\"u}nz, M.~Buffier, and L.~Agapito, ``{MaskFusion}: Real-time recognition, tracking and reconstruction of multiple moving objects,'' in \emph{IEEE International Symposium on Mixed and Augmented Reality (ISMAR)}.\hskip 1em plus 0.5em minus 0.4em\relax IEEE, 2018, pp. 10--20.

\bibitem[Cootes et~al.(1995)Cootes, Taylor, Cooper, and Graham]{Cootes95cviu}
T.~F. Cootes, C.~J. Taylor, D.~H. Cooper, and J.~Graham, ``Active shape models - their training and application,'' \emph{Comput. Vis. Image Underst.}, vol.~61, no.~1, pp. 38--59, January 1995.

\bibitem[Zhou et~al.(2015)Zhou, Leonardos, Hu, and Daniilidis]{Zhou15cvpr}
X.~Zhou, S.~Leonardos, X.~Hu, and K.~Daniilidis, ``{3D} shape reconstruction from {2D} landmarks: A convex formulation,'' in \emph{IEEE Conf. on Computer Vision and Pattern Recognition (CVPR)}, 2015.

\bibitem[Yang et~al.(2020{\natexlab{b}})Yang, Antonante, Tzoumas, and Carlone]{Yang20ral-GNC}
H.~Yang, P.~Antonante, V.~Tzoumas, and L.~Carlone, ``Graduated non-convexity for robust spatial perception: From non-minimal solvers to global outlier rejection,'' \emph{{IEEE} Robotics and Automation Letters ({RA-L})}, vol.~5, no.~2, pp. 1127--1134, 2020, arXiv preprint:1909.08605 (with supplemental material), \linkToPdf{https://arxiv.org/pdf/1909.08605.pdf}\award{, ICRA Best paper award in Robot Vision}.

\bibitem[Bar-Shalom(1992)]{BarShalom92}
Y.~Bar-Shalom, \emph{Multitarget multisensor tracking: {A}dvanced applications}.\hskip 1em plus 0.5em minus 0.4em\relax Norwood, MA: Artech House, 1992.

\bibitem[Blackman and Popoli(1999)]{Blackman99-modernTracking}
\BIBentryALTinterwordspacing
S.~Blackman and R.~Popoli, \emph{Design and Analysis of Modern Tracking Systems}, ser. Artech House radar library.\hskip 1em plus 0.5em minus 0.4em\relax Artech House, 1999. [Online]. Available: \url{https://books.google.com/books?id=lTIfAQAAIAAJ}
\BIBentrySTDinterwordspacing

\bibitem[Piga et~al.(2021)Piga, Bottarel, Fantacci, Vezzani, Pattacini, and Natale]{Piga2021frontiers-maskukf}
N.~A. Piga, F.~Bottarel, C.~Fantacci, G.~Vezzani, U.~Pattacini, and L.~Natale, ``Maskukf: An instance segmentation aided unscented kalman filter for 6d object pose and velocity tracking,'' \emph{Frontiers in Robotics and AI}, vol.~8, p. 594583, 2021.

\bibitem[Simon and Berger(1998)]{Simon98iccv-tracking}
G.~Simon and M.-O. Berger, ``A two-stage robust statistical method for temporal registration from features of various type,'' in \emph{Sixth International Conference on Computer Vision (IEEE Cat. No.98CH36271)}, 1998, pp. 261--266.

\bibitem[Simon and Berger(2002)]{Simon02-planarPose}
------, ``Pose estimation for planar structures,'' \emph{IEEE Computer Graphics and Applications}, vol.~22, no.~6, pp. 46--53, 2002.

\bibitem[Wen and Bekris(2021)]{Wen21iros-bundletrack}
B.~Wen and K.~Bekris, ``Bundletrack: 6d pose tracking for novel objects without instance or category-level 3d models,'' in \emph{IEEE/RSJ Intl. Conf. on Intelligent Robots and Systems (IROS)}.\hskip 1em plus 0.5em minus 0.4em\relax IEEE, 2021, pp. 8067--8074.

\bibitem[Yang and Carlone(2022)]{Yang22pami-certifiablePerception}
H.~Yang and L.~Carlone, ``Certifiably optimal outlier-robust geometric perception: Semidefinite relaxations and scalable global optimization,'' \emph{{IEEE} Trans. Pattern Anal. Machine Intell.}, 2022, \linkToPdf{https://arxiv.org/pdf/2109.03349.pdf}.

\bibitem[Lasserre(2010)]{Lasserre10book-momentsOpt}
J.~Lasserre, \emph{Moments, positive polynomials and their applications}.\hskip 1em plus 0.5em minus 0.4em\relax World Scientific, 2010, vol.~1.

\bibitem[Shor(1987)]{Shor87}
N.~Shor, ``Quadratic optimization problems,'' \emph{Izv. Akad. Nauk SSSR Tekhn. Kibernet.}, vol.~1, pp. 128--139, 1987.

\bibitem[Brynte et~al.(2021)Brynte, Larsson, Iglesias, Olsson, and Kahl]{Brynte21-semidefiniteRotation}
\BIBentryALTinterwordspacing
L.~Brynte, V.~Larsson, J.~P. Iglesias, C.~Olsson, and F.~Kahl, ``On the tightness of semidefinite relaxations for rotation estimation,'' \emph{Journal of Mathematical Imaging and Vision}, vol.~64, no.~1, p. 57–67, Oct. 2021. [Online]. Available: \url{http://dx.doi.org/10.1007/s10851-021-01054-y}
\BIBentrySTDinterwordspacing

\bibitem[Saunderson et~al.(2014)Saunderson, Parrilo, and Willsky]{Saunderson14cdc}
J.~Saunderson, P.~Parrilo, and A.~Willsky, ``Semidefinite relaxations for optimization problems over rotation matrices,'' in \emph{IEEE Conf. on Decision and Control (CDC)}, May 2014.

\bibitem[Rosen et~al.(2018)Rosen, Carlone, Bandeira, and Leonard]{Rosen18ijrr-sesync}
D.~Rosen, L.~Carlone, A.~Bandeira, and J.~Leonard, ``{SE-Sync}: a certifiably correct algorithm for synchronization over the {Special Euclidean} group,'' \emph{Intl. J. of Robotics Research}, 2018, arxiv preprint: 1611.00128, \linkToPdf{https://arxiv.org/abs/1611.00128}.

\bibitem[Carlone and Dellaert(2015)]{Carlone15icra-verification}
L.~Carlone and F.~Dellaert, ``Duality-based verification techniques for {2D SLAM},'' in \emph{IEEE Intl. Conf. on Robotics and Automation (ICRA)}, 2015, pp. 4589--4596

\bibitem[Briales and Gonzalez-Jimenez(2017)]{Briales17cvpr-registration}
J.~Briales and J.~Gonzalez-Jimenez, ``{Convex Global 3D Registration with Lagrangian Duality},'' in \emph{IEEE Conf. on Computer Vision and Pattern Recognition (CVPR)}, 2017.

\bibitem[Zhao et~al.(2020)Zhao, Xu, and Kneip]{Zhao20cvpr-certifiablyEssential}
J.~Zhao, W.~Xu, and L.~Kneip, ``A certifiably globally optimal solution to generalized essential matrix estimation,'' in \emph{IEEE Conf. on Computer Vision and Pattern Recognition (CVPR)}, 2020.

\bibitem[Garcia-Salguero et~al.(2021)Garcia-Salguero, Briales, and Gonzalez-Jimenez]{Garcia21IVC-certifiablerelativepose}
M.~Garcia-Salguero, J.~Briales, and J.~Gonzalez-Jimenez, ``Certifiable relative pose estimation,'' \emph{Image and Vision Computing}, vol. 109, p. 104142, 2021.

\bibitem[{Sun} and {Deng}(2020)]{Sun20access-certifiablyPnP}
L.~{Sun} and Z.~{Deng}, ``Certifiably optimal and robust camera pose estimation from points and lines,'' \emph{IEEE Access}, 2020.

\bibitem[Holmes et~al.(2024)Holmes, Dümbgen, and Barfoot]{Holmes24-semidefinite}
C.~Holmes, F.~Dümbgen, and T.~D. Barfoot, ``On semidefinite relaxations for matrix-weighted state-estimation problems in robotics,'' 2024.

\bibitem[Pavlakos et~al.(2017)Pavlakos, Zhou, Chan, Derpanis, and Daniilidis]{Pavlakos17icra-semanticKeypoints}
G.~Pavlakos, X.~Zhou, A.~Chan, K.~Derpanis, and K.~Daniilidis, ``6-dof object pose from semantic keypoints,'' in \emph{IEEE Intl. Conf. on Robotics and Automation (ICRA)}, 2017.

\bibitem[Tron et~al.(2015)Tron, Rosen, and Carlone]{Tron15rssws3D-dualityPGO3D}
R.~Tron, D.~Rosen, and L.~Carlone, ``On the inclusion of determinant constraints in lagrangian duality for {3D SLAM},'' in \emph{Robotics: Science and Systems (RSS), Workshop ``The problem of mobile sensors: Setting future goals and indicators of progress for {SLAM}''}, 2015, \linkToPdf{https://www.dropbox.com/s/859umrdf7ldd2kv/2015ws-rss-duality3Ddet.pdf?dl=0}.

\bibitem[Briales and Gonzalez-Jimenez(2016)]{Briales16iros}
J.~Briales and J.~Gonzalez-Jimenez, ``Fast global optimality verification in {3D SLAM},'' in \emph{IEEE/RSJ Intl. Conf. on Intelligent Robots and Systems (IROS)}, Oct 2016, pp. 4630--4636.

\bibitem[Eriksson et~al.(2018)Eriksson, Olsson, Kahl, and Chin]{Eriksson18cvpr-strongDuality}
A.~Eriksson, C.~Olsson, F.~Kahl, and T.-J. Chin, ``Rotation averaging and strong duality,'' \emph{IEEE Conf. on Computer Vision and Pattern Recognition (CVPR)}, 2018.

\bibitem[{MOSEK ApS}(2017)]{mosek}
\BIBentryALTinterwordspacing
{MOSEK ApS}, \emph{The MOSEK optimization toolbox for MATLAB manual. Version 8.1.}, 2017. [Online]. Available: \url{http://docs.mosek.com/8.1/toolbox/index.html}
\BIBentrySTDinterwordspacing

\bibitem[Shi et~al.(2021)Shi, Yang, and Carlone]{Shi21rss-pace}
J.~Shi, H.~Yang, and L.~Carlone, ``Optimal pose and shape estimation for category-level {3D} object perception,'' in \emph{Robotics: Science and Systems (RSS)}, 2021, arXiv preprint: 2104.08383, \linkToPdf{https://arxiv.org/pdf/2104.08383.pdf}, \linkToVideo{https://youtu.be/kiNBS0IF2-g}\award{, finalist for Best Paper Award}.

\bibitem[Ge et~al.(2022)Ge, Huangfu, Wang, Wu, and Ye]{copt}
D.~Ge, Q.~Huangfu, Z.~Wang, J.~Wu, and Y.~Ye, ``Cardinal {O}ptimizer {(COPT)} user guide,'' https://guide.coap.online/copt/en-doc, 2022.

\bibitem[Xiang et~al.(2014)Xiang, Mottaghi, and Savarese]{Xiang14-Pascal3DPlus}
Y.~Xiang, R.~Mottaghi, and S.~Savarese, ``Beyond pascal: A benchmark for 3d object detection in the wild,'' in \emph{IEEE Winter Conf. on Appl. of Computer Vision}.\hskip 1em plus 0.5em minus 0.4em\relax IEEE, 2014, pp. 75--82.

\bibitem[Calli et~al.(2015)Calli, Singh, Walsman, Srinivasa, Abbeel, and Dollar]{Calli15-YCBobject}
B.~Calli, A.~Singh, A.~Walsman, S.~Srinivasa, P.~Abbeel, and A.~M. Dollar, ``The {{YCB}} object and {{Model}} set: {{Towards}} common benchmarks for manipulation research,'' in \emph{Intl. Conf. on Advanced Robotics (ICAR)}, Jul. 2015, pp. 510--517.

\bibitem[He et~al.(2016)He, Zhang, Ren, and Sun]{He16cvpr-ResNet}
K.~He, X.~Zhang, S.~Ren, and J.~Sun, ``Deep residual learning for image recognition,'' pp. 770--778, 2016.

\bibitem[Denninger et~al.(2023)Denninger, Winkelbauer, Sundermeyer, Boerdijk, Knauer, Strobl, Humt, and Triebel]{Denninger23joss-blenderproc}
\BIBentryALTinterwordspacing
M.~Denninger, D.~Winkelbauer, M.~Sundermeyer, W.~Boerdijk, M.~Knauer, K.~H. Strobl, M.~Humt, and R.~Triebel, ``Blenderproc2: A procedural pipeline for photorealistic rendering,'' \emph{Journal of Open Source Software}, vol.~8, no.~82, p. 4901, 2023. [Online]. Available: \url{https://doi.org/10.21105/joss.04901}
\BIBentrySTDinterwordspacing

\bibitem[Xiang et~al.(2018)Xiang, Schmidt, Narayanan, and Fox]{Xiang17rss-posecnn}
Y.~Xiang, T.~Schmidt, V.~Narayanan, and D.~Fox, ``{PoseCNN}: A convolutional neural network for {6D} object pose estimation in cluttered scenes,'' in \emph{Robotics: Science and Systems (RSS)}, 2018.

\bibitem[Smit(2020)]{Smit20-ketchup}
\BIBentryALTinterwordspacing
A.~Smit, ``Heinz ketchup bottle,'' Oct 2020. [Online]. Available: \url{https://grabcad.com/library/heinz-ketchup-bottle-1}
\BIBentrySTDinterwordspacing

\bibitem{yolov8}Jocher, G., Chaurasia, A. \& Qiu, J. Ultralytics YOLO.  (2023,1), https://github.com/ultralytics/ultralytics

\bibitem[Deng et~al.(2022)Deng, Geng, Bretl, Xiang, and Fox]{Deng22ral-icaps}
X.~Deng, J.~Geng, T.~Bretl, Y.~Xiang, and F.~Dieter, ``iCaps: Iterative Category-level Object Pose and Shape Estimation,'' \emph{{IEEE} Robotics and Automation Letters ({RA-L})}, vol.~7, no.~2, pp. 1784--1791, 2022.

\bibitem[Ubellacker et~al.(2024)Ubellacker, Ray, Bern, Strader, and Carlone]{Ubellacker24npj-softDrone3}
S.~Ubellacker, A.~Ray, J.~Bern, J.~Strader, and L.~Carlone, ``High-speed aerial grasping using a soft drone with onboard perception,'' \emph{Nature Robotics}, 2024

\end{thebibliography}

}

\newpage
\normalsize
\appendices
\renewcommand{\theequation}{A\arabic{equation}}

\section{{Maximum A Posteriori Derivation}}
\label{appendix:optprob}
Here we show that \eqref{eq:optprob_original} is a maximum a posteriori (MAP) estimator. We first restate the problem:
\begin{equation}
    \label{eq:optprob_restated}
    \begin{aligned}[c]
        \min_{
            \substack{
                \BR_t,\BOmega_l\in \SO{3}, \\
                \Bp_t,\Bv_l\in\mathbb{R}^{3},\\
                \mathbf{c}\in\mathbb{R}^K,\mathbf{1}_K^\mathsf{T}\mathbf{c}=1\\
                t = 1,\hdots,T,\\
                l = 1,\hdots,T-1
                }
        }
        \\[3pt]
        \textrm{s.t.} & 
    \end{aligned}
    \begin{aligned}[c]
        &\sum_{t=1}^T \sum_{i=1}^N w_t^i \left\| \mathbf{y}_t^i - \BR_t\mathbf{B}_i\mathbf{c} - \mathbf{p}_t\right\|^2 
        + \lambda\|\Delta\mathbf{c}\|^2
        &&\\
        &+ \sum_{t=1}^{T-1} \omega_t\|\Delta\Bv_t\|^2
        + \kappa_t\|\Delta\BOmega_t\|^2_F &&\\
        &\mathbf{p}_{t+1} = \mathbf{p}_t + \mathbf{R}_t\mathbf{v}_t,\:\:
        \mathbf{R}_{t+1} = \mathbf{R}_t\boldsymbol{\Omega}_t&&\\
        &\Delta\mathbf{v}_t = \mathbf{v}_{t+1} - \mathbf{v}_t,\:\:
         \Delta\boldsymbol{\Omega}_t = \BOmega_{t+1}-\BOmega_t
        &&
    \end{aligned}
\end{equation}
where we introduced auxiliary variables $\Delta\mathbf{v}_t$ and $\Delta\boldsymbol{\Omega}_t$ for the velocity changes.
We now show that the first summand in~\eqref{eq:optprob_restated} corresponds to the likelihood of our keypoint measurements~\eqref{eq:measurements}, while the 
other terms describe our priors on the shape, velocity, and rotation rates. Denote the quantities to estimate by $\mathbf{z}\triangleq [\mathbf{c}, \{\mathbf{p}_t, \mathbf{R}_t\}_{t=1}^T, \{\mathbf{v}_t, \boldsymbol{\Omega}_t\}_{t=1}^{T-1}, \{\Delta\mathbf{v}_t, \Delta\boldsymbol{\Omega}_t\}_{t=1}^{T-1}]$ belonging to the domain $\mathbb{Z}$ which includes all relevant constraints in~\eqref{eq:optprob_restated}. The MAP estimator takes the form:
\begin{equation}
    \label{eq:map_full}
    \argmax_{z\in\mathbb{Z}}\ \mathbb{P}(\mathbf{z}\,|\,\{\mathbf{y}_t^i\}_{i,t=1}^{N,T})
    =
    \argmax_{z\in\mathbb{Z}}\ \mathbb{P}(\{\mathbf{y}_t^i\}_{i,t=1}^{N,T}\,|\,\mathbf{z})\mathbb{P}(\mathbf{z})
\end{equation}
where we expanded using Bayes rule. Assuming independent measurements, shape independence, and Markovian time-independence, we can rewrite \eqref{eq:map_full} as:
\begin{equation}
    \argmax_{z\in\mathbb{Z}}\:
    \prod_{i,t=1}^{N,T} \mathbb{P}(\mathbf{y}_t^i\,|\,\mathbf{z})
    \prod_{t=1}^{T-1} \mathbb{P}(\Delta\mathbf{v}_t)\mathbb{P}(\Delta\boldsymbol{\Omega}_t)
    \mathbb{P}(\Delta\mathbf{c})
\end{equation}
For the posterior $\mathbb{P}(\mathbf{y}_t^i\,|\,\mathbf{z})$ we assume a zero-mean Gaussian with covariance $\boldsymbol{\Sigma}_t^i = \frac{1}{w_t^i}\mathbf{I}_3$. Hence, using \eqref{eq:measurements} :
\begin{equation}
    \mathbb{P}(\mathbf{y}_t^i\,|\,\mathbf{z})
    =
    \alpha_t^i \exp\!\left( 
        -\frac{w_t^i}{2}\left\|
            \mathbf{y}_t^i - \BR_t\mathbf{B}_i\mathbf{c} - \mathbf{p}_t
        \right\|^2
    \right)
\end{equation}
with normalization constant $\alpha_t^i$.

Similarly, for velocity and shape we assume a zero-mean Gaussian prior with covariance $\frac{1}{\omega_t}\mathbf{I}_3$ and $\frac{1}{\lambda}\mathbf{I}_3$ respectively:
\begin{equation}
    \mathbb{P}(\Delta \mathbf{v}_t) = 
    \alpha_t^v \exp\!\left( 
        -\frac{\omega_t}{2}\left\|
            \Delta\mathbf{v}_t
        \right\|^2
    \right)
\end{equation}
\begin{equation}
    \mathbb{P}(\Delta\mathbf{c}) = 
    \alpha_c \exp\!\left( 
        -\frac{\lambda}{2}\left\|
            \Delta\mathbf{c}
        \right\|^2
    \right)
\end{equation}
We also assume that the rotation rate follows a Langevin distribution with concentration parameter $\kappa_t$:
\begin{equation}
    \mathbb{P}(\Delta\boldsymbol{\Omega}_t) = 
    \alpha_t^o \exp\!\left( 
        -\kappa_t\left\|
            \Delta\boldsymbol{\Omega}_t
        \right\|^2_F
    \right)
\end{equation}
where $\alpha_t^v$, $\alpha_c$, and $\alpha_t^o$ are suitable normalization constants.

Replacing the maximum of the posterior with the minimum of the negative logarithm of the posterior and dropping multiplicative and additive constants we arrive at the result.

\section{Proof of Proposition~\ref{prop:shapesoln}: \\ Closed-Form Optimal Shape}
\label{appendix:shape}
 Holding all other variables constant, \eqref{eq:optprob_original} is a linearly constrained least squares program in $\mathbf{c}$. Thus, the minimum with respect to $\mathbf{c}$ is convex and admits a unique solution via the KKT conditions. If we drop objective terms that do not depend on $\mathbf{c}$ in \eqref{eq:optprob_original}, we get:

\begin{equation}
    \begin{aligned}[c]
        \min_{
            \substack{
                \mathbf{c}\in\mathbb{R}^K,\\\mathbf{1}_K^\mathsf{T}\mathbf{c}=1
            }
        }
    \end{aligned}
    \quad
    \begin{aligned}[c]
        &\sum_{t=1}^T \sum_{i=1}^N w_t^i \left\| \mathbf{y}_t^i - \BR_t\mathbf{B}_i\mathbf{c} - \mathbf{p}_t\right\|^2 
        + \lambda\|\Delta\mathbf{c}\|^2&&
    \end{aligned}
\end{equation}

Expanding the summation over keypoint indices $i$ and moving the weights into the norm:
\begin{equation}
    \begin{aligned}[c]
        &\sum_{t=1}^T 
        \left\|\vphantom{
            \begin{bmatrix} 
                \sqrt{w_t^1}\mathbf{I}_3 & &\\
                & \ddots &\\
                &  & \sqrt{w_t^N}\mathbf{I}_3
            \end{bmatrix}
        }\right.
        \begingroup
        \setlength\arraycolsep{-5pt}
        \underbrace{
        \begin{bmatrix} 
            \sqrt{w_t^1}\mathbf{I}_3 & &\\
            & \ddots &\\
            &  & \sqrt{w_t^N}\mathbf{I}_3
        \end{bmatrix}
        }_{\triangleq \mathbf{W}_t}
        \endgroup
        \left(\vphantom{
            \begin{bmatrix}
                \mathbf{R}^\mathsf{T}_t ( \mathbf{y}^1_t - \mathbf{p}_t ) \\
                \vdots \\
                \mathbf{R}^\mathsf{T}_t (\mathbf{y}^N_t - \mathbf{p}_t )
            \end{bmatrix}
            - 
            \begin{bmatrix}
                \mathbf{B}_1\\
                \vdots \\
                \mathbf{B}_N
            \end{bmatrix}
            \mathbf{c}}
        \right.
            \underbrace{
            \begin{bmatrix}
                \mathbf{R}^\mathsf{T}_t ( \mathbf{y}^1_t - \mathbf{p}_t ) \\
                \vdots \\
                \mathbf{R}^\mathsf{T}_t (\mathbf{y}^N_t - \mathbf{p}_t )
            \end{bmatrix}
            }_{\triangleq\mathbf{h}_t}
            - 
            \underbrace{
            \begin{bmatrix}
                \mathbf{B}_1\\
                \vdots \\
                \mathbf{B}_N
            \end{bmatrix}
            }_{\triangleq\mathbf{B}}
            \mathbf{c}
        \left.\vphantom{
            \begin{bmatrix}
                \mathbf{R}^\mathsf{T}_t ( \mathbf{y}^1_t - \mathbf{p}_t ) \\
                \vdots \\
                \mathbf{R}^\mathsf{T}_t (\mathbf{y}^N_t - \mathbf{p}_t )
            \end{bmatrix}
            - 
            \begin{bmatrix}
                \mathbf{B}_1\\
                \vdots \\
                \mathbf{B}_N
            \end{bmatrix}
            \mathbf{c}}
        \right)
        \left.\vphantom{
            \begin{bmatrix} 
                \sqrt{w_t^1}\mathbf{I}_3 & &\\
                & \ddots &\\
                &  & \sqrt{w_t^N}\mathbf{I}_3
            \end{bmatrix}
        }\right\|
        ^2 \\&
        + \lambda\|\Delta\mathbf{c}\|^2&&
    \end{aligned}
\end{equation}
where we used the 2-norm rotational invariance to move $\mathbf{R}^\mathsf{T}$.

We can now write the stationarity condition. Using the dual variable $\mu$ for the condition $\mathbf{1}^\mathsf{T}\mathbf{c}=1$ and simplifying:
\begin{equation}
    0 = 2\mathbf{B}^\mathsf{T}\left(\sum_{t=1}^T W_t^2\right)\mathbf{B}\mathbf{c}
    - 2\mathbf{B}^\mathsf{T}\left(\sum_{t=1}^T W_t^2 \mathbf{h}_t\right) + 2\lambda\Delta\mathbf{c} + \mathbf{1}\mu
\end{equation}

Putting this together with primal feasibility, we arrive at the following linear system:
\begin{equation}
    \begin{bmatrix}
        \mathbf{H}^{-1} & \mathbf{1}_K \\
        \mathbf{1}_K^\mathsf{T} & 0
    \end{bmatrix}
    \begin{bmatrix}
        \mathbf{c} \\ \mu
    \end{bmatrix} \\
    =
    \begin{bmatrix}
        2\left(\mathbf{B}^\mathsf{T}\sum_{t=1}^T \mathbf{W}_t^2 \mathbf{h}_t + \lambda \mathbf{\bar{c}}\right) \\
        1
    \end{bmatrix}
\end{equation}
where $\mathbf{H} \triangleq \frac{1}{2}\left(\mathbf{B}^\mathsf{T}\left(\sum_{t=1}^T \mathbf{W}_t^2\right)\mathbf{B} + \lambda\mathbf{I}_K\right)^{-1}$.

Solving for $\mathbf{c}$ and substituting the definitions of $\mathbf{G}$ and $\mathbf{g}$, we arrive at the result. Cruicially, notice that the matrix we must invert to get $\mathbf{H}$ is made up only of constants.

\section{Proof of Proposition~\ref{prop:qcqpform}: \\Quadratically Constrained Quadratic Program}
\label{appendix:quartic2qcqp}
We focus on the measurement terms, the constraints, and the variable $\mathbf{c}$. The remaining objective terms contain norms of single-degree variables and are thus quadratic. The key idea is to let $\mathbf{s}_t\triangleq\mathbf{R}_t^\mathsf{T}\mathbf{p}_t$. Then, the measurement term of the objective may be rotated without changing its norm:
\begin{equation}
    \left\| \mathbf{y}_t^i - \BR_t\mathbf{B}_i\mathbf{c} - \mathbf{p}_t\right\|^2 = 
    \left\| \BR_t^\mathsf{T}\mathbf{y}_t^i - \mathbf{B}_i\mathbf{c} - \mathbf{s}_t\right\|^2
\end{equation}

Similarly, the optimal solution $\mathbf{c^*}$ may be rewritten as:
\begin{equation}
    \mathbf{c^\star} = 2\mathbf{G}\left(
            \mathbf{B}^\mathsf{T}
            \sum_{t=1}^T 
            \mathbf{W}_t
            \begin{bmatrix}
                \mathbf{R}^\mathsf{T}_t \mathbf{y}^1_t - \mathbf{s}_t \\
                \vdots \\
                \mathbf{R}^\mathsf{T}_t \mathbf{y}^N_t - \mathbf{s}_t
            \end{bmatrix}
            + \lambda \mathbf{\bar{c}}\right) + \mathbf{g}
\end{equation}
to complete the changes needed for the objective. Notice that $\mathbf{c}$ is a linear function of $\mathbf{R}$ and $\mathbf{s}$. Thus, every term of the objective is quadratic in the new variables $\mathbf{s}_t$, $\mathbf{R}_t$, $\mathbf{v}_t$, $\boldsymbol{\Omega}_t$.

\begin{figure*}[htb!]
    \vspace{-10pt}
    \centerline{\includegraphics[width=\linewidth]{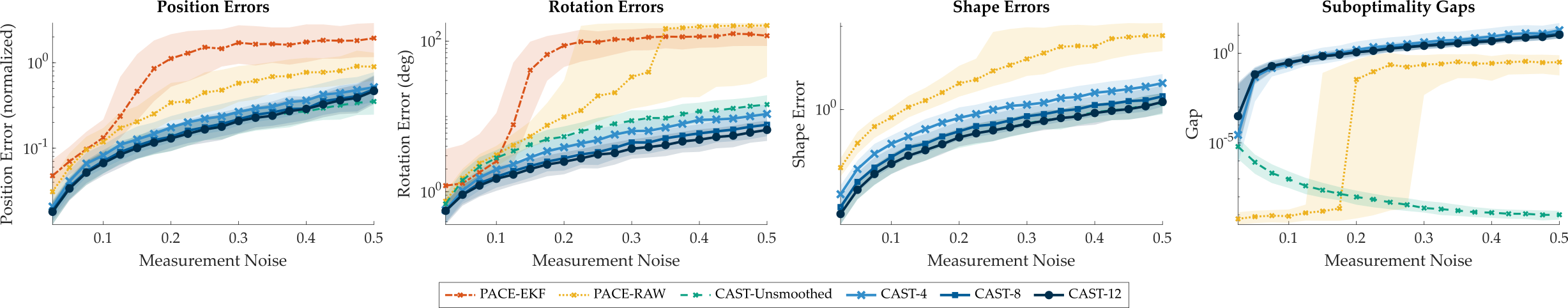}}
    \caption{\textbf{Performance of \name in synthetic experiments with increasing measurement noise.} Robustness to measurement noise with \name using the inverse of the simulated velocity covariance for the velocity weights $\omega_t$. The key difference between this plot and \prettyref{fig:synthetic_results}(a) lies in the suboptimality gap figure, where \name loses tightness quickly. Despite losing its optimality certificate, \name maintains the lowest position, rotation, and shape errors.}
    \label{fig:mle_mnoise}
    \vspace{-10pt}
\end{figure*}

The variable $\mathbf{p}$ still remains in the constraint $\mathbf{p}_{t+1} = \mathbf{p}_t + \mathbf{R}_t\mathbf{v}_t$. Multiplying both sides by $\mathbf{R}_t^\mathsf{T}$:
\begin{equation}
    \mathbf{R}_t^\mathsf{T}\mathbf{p}_{t+1} = \mathbf{s}_t + \mathbf{v}_t
\end{equation}
From the rotation rate constraint, $\mathbf{R}_{t+1} = \mathbf{R}_t \boldsymbol{\Omega}_t \Rightarrow \mathbf{R}_t^\mathsf{T} = \boldsymbol{\Omega}_t \mathbf{R}_{t+1}^\mathsf{T}$. Plugging this in gives the desired constraint.

\section{Proof of \prettyref{prop:timecompatibility}: \\ Time Compatibility Test}
\label{appendix:timecompproof}
    Compare the distance between keypoints $i$ and $j$ at each time, rotating to align coordinates with the body frame:
    \begin{equation}
        \label{eq:timecomp_initial}
        \left|\|\mathbf{R}_l^\mathsf{T}(\mathbf{y}_l^i - \mathbf{y}_l^j))\| - \|\mathbf{R}_m^\mathsf{T}(\mathbf{y}_m^i - \mathbf{y}_m^j)\|\right|
    \end{equation}
    Bound this using the {reverse and forward triangle} inequalities, noting that noise is isotropic ($\mathbf{R}\boldsymbol{\epsilon}=\boldsymbol{\epsilon}$):
    \begin{equation}
        \label{eq:timecomp_revtriangle}
        \begin{aligned}
            \eqref{eq:timecomp_initial}& {\leq}
            \|(\mathbf{B}_i - \mathbf{B}_j + \boldsymbol{\epsilon}_l^i - \boldsymbol{\epsilon}_l^j) - 
            (\mathbf{B}_i - \mathbf{B}_j + \boldsymbol{\epsilon}_m^i - \boldsymbol{\epsilon}_m^j)\|\\
            &= \|\boldsymbol{\epsilon}_y^{i,l} - \boldsymbol{\epsilon}_y^{i,l}
            - \boldsymbol{\epsilon}_y^{i,m} + \boldsymbol{\epsilon}_y^{i,m}\| \leq 4\epsilon
        \end{aligned}
    \end{equation}
    Since the 2-norm is invariant to rotations, we can remove the rotations from \eqref{eq:timecomp_initial} and  obtain the result.

\section{Additional Experimental Results}
\label{appendix:exp_results}
\begin{figure}[htb!]
    \centerline{\includegraphics[width=\linewidth]{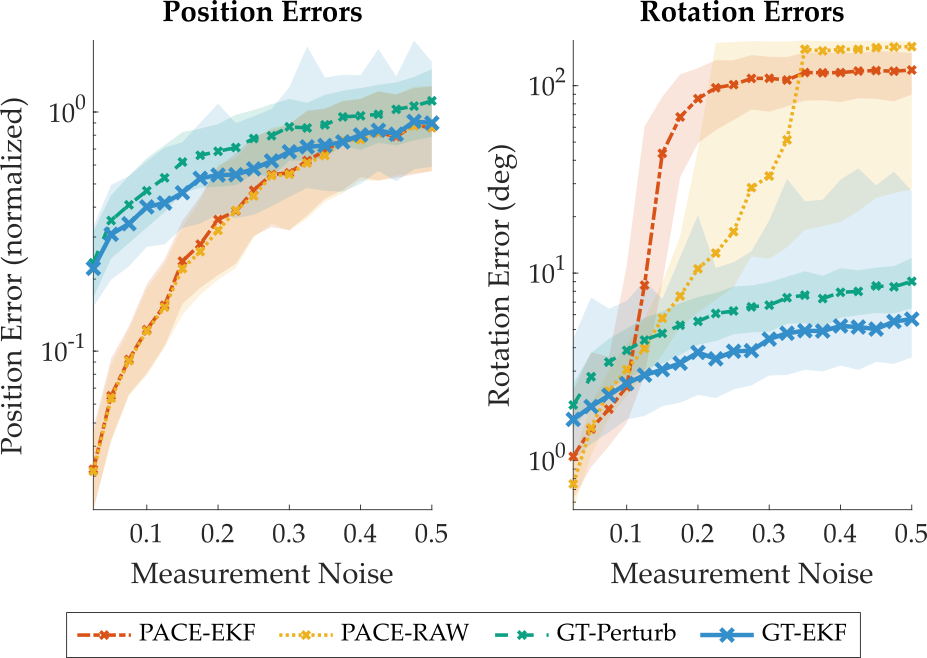}}
    \caption{\textbf{Extended Kalman Filter with perturbed ground truth measurements.} 
    With Gaussian-perturbed ground truth measurements, the extended Kalman filter outperforms the raw measurements in median error across measurement noise values. This supports our claim that the EKF performs poorly using pose estimate from PACE, likely due to the high variance and heavy-tailed distribution of the estimates.}
    \label{fig:appendix_EKF}
\end{figure}

\textbf{Additional Synthetic Results.} In Section~\ref{sec:exp:synthetic} we showed the robustness of \name to measurement and process noise, and the robustness of \nameSharp to outliers. Here we give the runtimes of each method (\prettyref{tab:appendix_runtimes}), show results for the choice of weights resulting from MAP estimation (\prettyref{fig:mle_mnoise}), and address the EKF results (\prettyref{fig:appendix_EKF}). 

\begin{table}[htb!]
    \centering
    \caption{Synthetic Experiment Runtimes}
    \label{tab:appendix_runtimes}
    \begin{tabular}{c|c|cccc}
        \hline
        \multirow{2}{*}{\textbf{Runtime (s)}} & \multirow{2}{*}{PACE}       & \multicolumn{4}{c}{CAST-}                                                                                   \\
                                              &                             & 4                         & 8                        & 12                       & U                        \\ \hline
        Meas. Noise                           & 0.0028                      & 0.483                     & 2.15                     & 5.49                     & 5.25                     \\
        Proc. Noise                           & \multicolumn{1}{l|}{0.0040} & \multicolumn{1}{l}{0.857} & \multicolumn{1}{l}{4.05} & \multicolumn{1}{l}{10.6} & \multicolumn{1}{l}{10.2}
        \\
        \hline
    \end{tabular}
\end{table}

From \prettyref{tab:appendix_runtimes} we observe \name is the slowest of the tested methods. 
The results are obtained with a non-optimized MATLAB implementation and we expect computational gains from further code optimization.
This aside, the variable horizon length allows a trade-off between computational speed and accuracy. As computation improves, the benefits of certifiable optimality and increased accuracy make \name an attractive choice of tracking algorithm.

Recall that in the tests in Section~\ref{sec:experiments} we chose the velocity weights to be $\omega_t = 1$ instead of setting them as prescribed by MAP estimation (where they should be taken as the inverse of the variance of the prior). 
This is equivalent to increasing the standard deviation of the velocity noise by a factor of 10; in other words, it reduces the effect of motion smoothing. \prettyref{fig:mle_mnoise} shows that using the true velocity covariance degrades tightness, although it does not have any visible effect on the accuracy results. 

Lastly, we present additional results showing the performance of the EKF using perturbed ground truth data instead of PACE. Specifically, we perturb the ground truth poses according to a zero-mean Gaussian with standard deviation equal to $1/25$th of the measurement noise for position and $1/50$th for rotation (arbitrarily chosen as realistic values). \prettyref{fig:appendix_EKF} shows the median EKF estimate consistently beats the perturbed ground truth value (results are averaged over 500 independent trials for each noise value). The large interquartile range is likely because of errors due to linearization, particularly of the constant twist motion model. The EKF likely struggled when using PACE's poses in the measurement update because of the high variance and heavy-tailed distribution of the estimates.

\begin{table}[htbp]
    \centering
    \caption{Additional YCBInEOAT Results}
    \label{tab:ycbineaot_soupbleach}
    \begin{tabular}{c|cc|cc}
    \hline
    \multirow{2}{*}{\textbf{Method}} & \multicolumn{2}{c|}{Bleach}                          & \multicolumn{2}{c}{Soup}                                        \\ 
                            & \multicolumn{1}{l}{ADD} & \multicolumn{1}{l|}{ADD-S} & \multicolumn{1}{l}{ADD}             & \multicolumn{1}{l}{ADD-S} \\ \hline
    6-PACK & 4.18                    & 18.00                      &12.82          & 60.32                      \\
    TEASER++      & 35.39                   & 46.40                      & 65.85         & 81.53                     \\
    MaskFusion    & 29.83                   & 43.31                      & {5.65}           & 6.45                       \\
    BundleTrack & \textbf{89.34}          & \textbf{94.72}             & {\textbf{86.00}} & 95.13           \\
    BundleSDF & 85.59                   & 93.11                     & 80.54        & \textbf{96.47}                      \\
    \nameSharp-8  & 47.53                   & 45.82                      &27.61              & 41.70                          \\ \hhline{=|==|==}
    \nameSharp-GT    & 62.19                   & 75.14                      & {37.07}              & 63.29                          \\ \hline
    \end{tabular}
\end{table}

\textbf{Results for Bleach and Soup on YCBInEOAT.} {Table~\ref{tab:ycbineaot_soupbleach} shows scores for all tested methods on the ``soup'' and ``bleach'' objects. As mentioned in the text, the soup object is particularly difficult because it is very small and cylindrically symmetric, which \name is not designed to handle (other approaches also achieve low scores, compared to the other objects). The bleach object is larger but matches the background color, making keypoint detection difficult.} %

\end{document}